\documentclass[preprint,12pt,3p]{elsarticle}

\usepackage{amssymb}
\usepackage[utf8]{inputenc}
\usepackage{threeparttable, tablefootnote}
\usepackage{footnote}
\usepackage{graphicx}
\usepackage{caption}
\usepackage{subcaption}
\usepackage{hyperref}
\usepackage{color}

\usepackage{amsmath}
\DeclareMathOperator*{\argmax}{arg\,max}
\usepackage{algorithm2e}
\usepackage{multirow}
\usepackage{floatrow}
\usepackage{hyperref}

\newcommand\Tstrut{\rule{0pt}{2.6ex}}         
\newcommand\Bstrut{\rule[-0.9ex]{0pt}{0pt}}   

\journal{ }

\begin{document}

\begin{frontmatter}

\title{Modeling Stated Preference for Mobility-on-Demand Transit: \\ A Comparison of Machine Learning and Logit Models}

\author[label2]{Xilei Zhao}
\address[label2]{H. Milton Stewart School of Industrial and Systems Engineering, Georgia Institute of Technology}
\ead{xilei.zhao@isye.gatech.edu}

\author[label5]{Xiang Yan\corref{cor1}}
\address[label5]{Taubman College of Architecture and Urban Planning, University of Michigan}
\ead{jacobyan@umich.edu}

\author[label1]{Alan Yu}
\address[label1]{Department of Electrical Engineering and Computer Science, University of Michigan}
\ead{alayu@umich.edu}

\cortext[cor1]{Corresponding Author.}

\author[label2]{Pascal Van Hentenryck}
\ead{pvh@isye.gatech.edu}

\begin{abstract}

Logit models are usually applied when studying individual travel behavior, i.e., to predict travel mode choice and to gain behavioral insights on traveler preferences. Recently, some studies have applied machine learning to model travel mode choice and reported higher out-of-sample predictive accuracy than traditional logit models (e.g., multinomial logit). However, little research focuses on comparing the interpretability of machine learning with logit models. In other words, how to draw behavioral insights from the high-performance ``black-box" machine-learning models remains largely unsolved in the field of travel behavior modeling.

This paper aims at providing a comprehensive comparison between the two approaches by examining the key similarities and differences in model development, evaluation, and behavioral interpretation between logit and machine-learning models for travel mode choice modeling. To complement the theoretical discussions, the paper also empirically evaluates the two approaches on the stated-preference survey data for a new type of transit system integrating high-frequency fixed-route services and ridesourcing. The results show that machine learning can produce significantly higher predictive accuracy than logit models. Moreover, machine learning and logit models largely agree on many aspects of behavioral interpretations. In addition, machine learning can automatically capture the nonlinear relationship between the input features and choice outcomes. The paper concludes that there is great potential in merging ideas from machine learning and conventional statistical methods to develop refined models for travel behavior research and suggests some new research directions.



\end{abstract}


\end{frontmatter}


\section{Introduction}
\label{sec1}

%



Emerging shared mobility services, such as car sharing, bike sharing, ridesouring, and micro-transit, have rapidly gained popularity across cities and are gradually changing how people move around. Predicting individual preferences for these services and the induced changes in  travel behavior is critical for transportation planning. Traditionally, travel behavior research has been primarily supported by discrete choice models (a type of statistical models), most notably the logit family such as the multinomial logit (MNL), the nested logit model and the mixed logit model. In recent years, as machine learning has become pervasive in many fields, there has been a growing interest in its application to model individual choice behavior. 

Machine learning and conventional statistical models seek to understand the data structure based on different approaches. The logit models, like many other statistical models, are based on a theoretical foundation which is mathematically proven, but this requires the input data to satisfy strong assumptions such as the random utility maximization decision rule and a particular type of error-term distribution \citep{ben1985discrete}. On the other hand, machine learning relies on computers to probe the data for its structure, without a theory of what the underlying data structure should look like. In other words, while a logit model presupposes a certain type of structure of the data with its behavioral and statistical assumptions, machine learning, on the other hand, ``lets the data speak for itself" and hence allows forming more flexible model structures, which can often lead to higher predictive capability (i.e., higher out-of-sample predictive accuracy). 


A number of recent empirical studies have verified that machine learning can outperform logit models in terms of predictive capability \citep[e.g.][]{xie2003work,zhang2008travel,hagenauer2017comparative,wang2018machine,lheritier2018airline}. Intuitively, a machine-learning model that predicts well should also be able to offer good interpretation by accurately representing the underlying data structure. However, existing studies that apply machine learning for travel-mode choice modeling have mostly focused on prediction, with much less attention being devoted to interpretation. More specifically, these studies rarely apply interpretable machine learning tools such as partial dependent plots and variable importance to extract behavioral findings from machine learning and compare/validate these findings with those obtained from traditional logit models. In mode-choice modeling applications, however, the behavioral interpretation of the results is as important as the prediction problem, since it offers valuable insights for transportation planning and policy making.

Furthermore, the existing literature comparing logit models and machine learning for modeling travel mode choice has two other limitations. First, the comparisons were usually made between the MNL model, the simplest logit model, and machine-learning algorithms of different complexity. In cases where the assumption of independence of irrelevant alternatives (IIA) is violated, such as when panel data (i.e., data containing multiple mode choices made by the same individuals) are examined, more advanced logit models such as the mixed logit model should be considered. 
Second, existing studies rarely discuss the fundamental differences in the application of machine-learning methods and logit models to travel mode choice modeling. The notable differences between the two approaches in the input data structure and data needs, the modeling of alternative-specific attributes, and the form of predicted outputs carry significant implications for model comparison. These differences and their implications, although touched on by some researchers such as \citet{omrani2013prediction}, have not been thoroughly examined.



This paper tries to bridge these gaps: It provides a comprehensive comparison of logit models and machine learning in modeling travel mode choices and also an empirical evaluation of the two approaches based on stated-preference (SP) survey data on a proposed mobility-on-demand transit system, i.e., an integrated transit system that runs high-frequency buses along major corridors and operates on-demand shuttles in the surrounding areas \citep{TS2017}. The paper first discusses the fundamental differences in the practical applications of the two types of methods, with a particular focus on the implications of the predictive performance of each approach and their capabilities to facilitate behavioral interpretations. The paper then compares the performance of two logit models (MNL and mixed logit) and seven machine-learning classifiers, including Naive Bayes (NB), classification and regression trees (CART), boosting trees (BOOST), bagging trees (BAG), random forest (RF), SVM, and NN, in predicting individual choices of four travel modes and their respective market shares. Moreover, the paper compares behavioral interpretations of two logit models (MNL and mixed logit) and two machine-learning models (RF and NN). The results show that RF can produce higher out-of-sample prediction accuracy than logit models and NN. Moreover, machine learning can offer consistent behavioral interpretations compared to logit models. In particular, RF can automatically capture nonlinearities between the input data and the choice outcome.

The rest of the paper is organized as follows. The next section provides a brief review of the literature in modeling mode choices using logit and machine-learning models. Section 3 explains the fundamentals of the logit and machine-learning models, including model formulation and input data structures, model development and evaluation, and model interpretation and application. Section 4 introduces the data used for empirical evaluation and Section 5 describes the logit and machine-learning models examined and their specifications. Section 6 evaluates these models in terms of predictive capability and interpretability. Lastly, Section 7 concludes by summarizing the findings, identifying the limitations of the paper, and suggesting future research directions. Table \ref{tab:acy} presents the list of abbreviations and acronyms used in this paper.

\begin{table}[!h]
    \centering
    \caption{List of Abbreviations and Acronyms.}
    \begin{tabular}{c|c}
    \hline
       MNL  &  Multinomial logit \\ \hline
       NB &  Naive Bayes \\ \hline
       CART & Classification and regression trees \\ \hline
       RF & Random forest \\\hline
       BOOST & Boosting trees \\\hline
       BAG & Bagging trees \\\hline
       SVM & Support vector machines \\\hline
       NN & Neural networks \\\hline
       AIC & Akaike information criterion \\\hline
       BIC & Bayesian information criterion \\\hline
       Min & Minimum \\\hline
       Max & Maximum \\\hline
       SD & Standard deviation \\\hline
       SP & Stated-preference \\\hline
       RP & Revealed-preference \\\hline
       IIA & Independence of irrelevant alternatives \\ \hline
       PT & Public transit \\ \hline
    \end{tabular}
    \label{tab:acy}
\end{table}

\section{Literature Review}
\label{sec2}

The logit family is a class of econometric models based on random utility maximization \citep{ben1985discrete}. Due to their statistical foundations and their capability to represent individual choice behavior realistically, the MNL model and its extensions have dominated travel behavior research ever since its formulation in the 1970s \citep{mcfadden1973conditional}. The MNL model is frequently challenged for its major assumption, the IIA property, and its inability to account for taste variations among different individuals. To address these limitations, researchers have developed important extensions to the MNL model such as the nested logit model and more recently the mixed logit model. The mixed logit model, in particular, has received much interest in recent years: Unlike the MNL model, it does not require the IIA assumption, can accommodate preference heterogeneity, and may significantly improve the MNL behavioral realism in representing consumer choices \citep{hensher2003mixed}.

Mode-choice modeling can also be viewed as a \textit{classification} problem, providing an alternative to logit models. A number of recent publications have suggested that machine-learning classifiers such as CART, NN, and SVM are effective in modeling individual travel behavior \citep{xie2003work,zhang2008travel,omrani2013prediction,omrani2015predicting,hagenauer2017comparative, golshani2018modeling, wang2018machine,wong2018discriminative,lheritier2018airline}. These studies generally found that machine-learning classifiers outperform traditional logit models in predicting travel-mode choices. For example, \citet{xie2003work} applied CART and NN to model travel mode choices for commuting trips taken by residents in the San Francisco Bay area. These machine-learning methods exhibited better performance than the MNL model in terms of prediction. Based on data collected in the same area, \citet{zhang2008travel} reported that SVM can predict commuter travel mode choice more accurately than NN and MNL. More recently, \citet{lheritier2018airline} found that the RF model outperforms the standard and the latent class MNL model in terms of accuracy and computation time, with less modeling effort.

It is not surprising that machine-learning classifiers can perform better than logit models in predictive tasks. Unlike logit models that make strong mathematical assumptions (i.e. constraining the model structure and assuming a certain distribution in the error term a priori), machine learning allows for more flexible model structures, which can reduce the model's incompatibility with the empirical data \citep{xie2003work,christopher2016pattern}. More fundamentally, the development of machine learning prioritizes predictive power, whereas advances in logit models are mostly driven by refining model assumptions, improving model fit, and enhancing the behavioral interpretation of the model results \citep{brownstone1998forecasting,hensher2003mixed}. In other words, the development of logit models prioritizes parameter estimation (i.e. obtaining better model parameter estimates that underline the relationship between the input features and the output variable) and pay less attention to increasing the model's out-of-sample predictive accuracy \citep{mullainathan2017machine}. In fact, recent studies have shown that the mixed logit model, despite resulting in substantial improvements in overall model fit, often resulted in poorer prediction accuracy compared to the simpler and more restrictive MNL model \citep{cherchi2010validation}.

While recognizing the superior predictive power of machine-learning models, researchers often think that they have weak explanatory power \citep{mullainathan2017machine}. In other words, machine-learning models are often regarded as ``not interpretable.'' Machine-learning studies rarely apply model outputs to facilitate behavioral interpretations, i.e., to test the response of the output variable or to changes in the input variables in order to generate findings on individual travel behavioral and preferences \citep{karlaftis2011statistical}. The outputs of many machine-learning models are indeed not directly interpretable as one may need hundreds of parameters to describe a deep NN or hundreds of decision trees to understand a RF model. Nonetheless, with recent development in interpretable/explainable machine learning, a wide range of machine learning interpretation tools have been invented to extract knowledge from the black-box models to facilitate decision-making \citep{golshani2018modeling,wager2018estimation,molnar2018interpretable}. In particular, \citet{zhao2019modeling} applied interpretable machine learning to model and assess heterogeneous travel behavior.
Examining these behavioral outputs from machine learning models could shed light on what factors are driving prediction decisions and also the fundamental question of whether machine learning is appropriate for behavioral analysis.


Prediction and behavioral analysis are equally important in travel behavior studies. While the primary goal of some applications is to accurately predict mode choices (and investigators are usually more concerned about the prediction of aggregate market share for each mode than about the prediction of individual choices), other studies may be more interested in quantifying the impact of different trip attributes on travel mode choices. To our knowledge, mode-choice applications that focus on behavioral outputs such as elasticity, marginal effect, value of time, and willingness-to-pay measures have received even more attention than those that focus on predicting individual mode choice or aggregate market shares in the literature. This paper thus extends current literature by comparing the behavioral findings from logit models and machine-learning methods, beyond the existing studies that primarily focus on their predictive accuracy.

Finally, this paper points out other differences in the practical applications of these two approaches that have bearings on model outputs and performance, including their input data structure and data needs, the treatment of alternative-specific attributes, and the forms of the predicted outputs. Discussions of these differences are largely absent from the current literature that compares the application of logit models and machine-learning algorithms in travel behavior research.

\section{Fundamentals of the Logit and Machine-Learning Models}

This section discusses the fundamentals of the logit and machine-learning models. Table \ref{tab:symbol_des} presents the list of symbols and notations used in the paper and Table \ref{tab:comparison} summarizes the comparison between logit and machine-learning models from various angles. The rest of this section describes this comparison in detail. 

\begin{table}[!t]
    \centering
    \caption{List of Symbols and Notations Used in the Paper}
    \footnotesize
    \begin{tabular}{c|c}
    \hline
     \bf{} Symbols    & \bf{} Description \\
     \hline
     $K$ & Total number of alternatives  \\
     \hline
     $N$ & Total number of observations \\
     \hline
     $P$ & Total number of features\\
     \hline
     $\boldsymbol{X}$  &  Input data for logit models containing $P$ features with $N$ observations for $K$ alternatives \\
     \hline
     $\boldsymbol X_{k,p}$ & Feature $p$ for alternative $k, k = 1, ..., K$ of $\boldsymbol X$\\
     \hline
     $\boldsymbol{X}_{k,-p}$ & All the features except $p$ for alternative $k, k = 1, ..., K$ of $\boldsymbol X$\\
     \hline
     $\boldsymbol{X}_{ik}$ & A row-vector for the $i$th observation for alternative $k, k = 1, ..., K$ \\
     \hline
     $\boldsymbol{X}_k$ & Input data for alternative $k$, $\boldsymbol{X}_k = [\boldsymbol{X}_{.k1}; ...; \boldsymbol{X}_{.kP}]$ where $\boldsymbol{X}_{.kp} = [X_{1kp}, ..., X_{Nkp}]$\\
     \hline
     $\boldsymbol{X}_i$  & The $i$th observation of $\boldsymbol{X}$, $\boldsymbol{X}_i = [\boldsymbol{X}_{i.1}, ..., \boldsymbol{X}_{i.P}]$ where $\boldsymbol{X}_{i.p} = [X_{i1p}; ...; X_{iKp}]$  \\
     \hline
     $\boldsymbol{X}_p$  &  The feature $p$ of $\boldsymbol{X}$,  $\boldsymbol{X}_p = [\boldsymbol{X}_{1.p}, ..., \boldsymbol{X}_{N.p}]$ where $\boldsymbol{X}_{i.p} = [X_{i1p}; ...; X_{iKp}]$\\
     \hline
     $\boldsymbol{Z}$  &  Input data for machine-learning models containing $P$ features and $N$ observations\\
     \hline
     $\boldsymbol Z_{p}$ & Feature $p$ of $\boldsymbol{Z}$\\
     \hline
     $\boldsymbol Z_{-p}$ & All the features except $p$ of $\boldsymbol{Z}$\\
     \hline
     $\boldsymbol{Z}_i$  &  $i$th observation of $\boldsymbol{Z}, \boldsymbol{Z}_i = [Z_{i1}, ..., Z_{iP}$]\\
     \hline
     $U_k(\boldsymbol{X}_k|\boldsymbol{\beta}_k)$ & Utility function for mode $k$ \\
     \hline
     $\boldsymbol{\beta}_k$ & Parameter vector for alternative $k$ of MNL model\\
     \hline
     $\boldsymbol{\beta}$ & Parameter matrix of MNL model, $\boldsymbol{\beta} = [\boldsymbol{\beta}_1,..., \boldsymbol{\beta}_K]$ \\
     \hline
     $\hat{\boldsymbol{\beta}}$ & Estimated parameter matrix of MNL model \\
     \hline
     $\boldsymbol{\varepsilon}_{k}$ & Random error for alternative $k$ of MNL model \\
     \hline
     $\boldsymbol{Y}$   & Output mode choice data \\
     \hline
     $\hat{Y}_i$   & Estimated mode choice for observation $i$ \\
     \hline
     $\boldsymbol{\theta}$   & Parameter or hyperparameter vector for machine-learning models\\
     \hline
     $\hat{\boldsymbol{\theta}}$   & Estimated parameter or hyperparameter vector\\
     \hline
     $f(\boldsymbol{Z}|\boldsymbol{\theta})$ & Machine-learning models based on $\boldsymbol{Z}$ and $\boldsymbol{\theta}$ \\
     \hline
     $p_{ik}$ & Probability of choosing alternative $k$ of observation $i$ \\
     \hline
     $\hat{p}_{ik}$ & Predicted probability for choosing alternative $k$ of observation $i$\\
     \hline
     $I_k(\hat{Y}_i)$ & Indicator function that equals to 1 if $\hat{Y}_i = k$ \\
     \hline
     $P_k(\boldsymbol{X}|\hat{ \boldsymbol{\beta}})$ & Aggregate level prediction for mode $k$ based on $\boldsymbol{X}$ and $\hat{ \boldsymbol{\beta}}$ for logit models\\
     \hline
     $Q_k(\boldsymbol{Z}|\hat{ \boldsymbol{\theta}})$ & Aggregate level prediction for mode $k$ based on $\boldsymbol{Z}$ and $\hat{ \boldsymbol{\theta}}$ for machine-learning models\\
     \hline
     $E_k(\cdot)$ & Arc elasticity for alternative $k$\\
     \hline
     $M_k(\cdot)$ & Marginal effect for alternative $k$\\
     \hline
     $\Delta$ & Constant\\
     \hline
    \end{tabular}
    \label{tab:symbol_des}
\end{table}

\begin{table}[!]
\caption{Comparison Between Logit and Machine-Learning Models}
\footnotesize
\resizebox{1\textwidth}{!}{
\begin{tabular}{p{3.2cm}|p{8.8cm}|p{7.5cm}}
\hline
\textbf{}                                      & \textbf{Logit Models}                           & \textbf{Machine-Learning Models}                                                                    \Tstrut\Bstrut               \\ \hline 
\multirow{3}{*}{\textbf{Model formulation}}   & $U_{k}(\boldsymbol{X}_k|\boldsymbol{\beta}_k) = \boldsymbol{\beta}_k^T \boldsymbol{X}_{k} + \boldsymbol{\varepsilon}_{k}$                       & $\boldsymbol{Y} = f(\boldsymbol{Z}|\boldsymbol{\theta}), \boldsymbol{Y} \in \{1, …, K\}$      \Tstrut                                                                           \\
                                               & $p_{ik} = \frac{\exp {(\boldsymbol{\beta}}_k^T \boldsymbol{X}_{ik})}{\sum_{p=1}^K \exp {(\boldsymbol{\beta}}_p^T \boldsymbol{X}_{ik})}, k \in \{1, ..., K\}$ \Tstrut                                            &                             
                                               \\ \hline
\textbf{Commonly used model type}              & MNL, mixed logit, nested MNL, generalized MNL   & NB, CART, BAG, BOOST, RF, SVM, NN                                                                 \Tstrut\Bstrut                  \\ \hline
\textbf{Prediction type}                       & Class probability: $p_{i1}, …, p_{iK}$                & Classification: $k, k \in \{1, ..., K\}$                                                \Tstrut\Bstrut                              \\ \hline
\textbf{Input data}                            & $\boldsymbol{X}$                 & $\boldsymbol{Z}$                                                                                                               \\ \hline
\textbf{Model topology}                        & Layer structure                                 & Layer structure, tree structure, case-based reasoning, etc.                           \Tstrut\Bstrut                              \\ \hline
\textbf{Optimization method}                   & Maximum likelihood estimation, simulated maximum likelihood                               & Back propagation, gradient descent, recursive partitioning, structural risk minimization, maximum likelihood, etc.    \\ \hline
\textbf{Evaluation criteria}                   & (Adjusted) McFadden's pseudo $R^2$, AIC, BIC & Resampling-based measures, e.g., cross validation                                                           \Tstrut\Bstrut                \\ \hline
\textbf{Individual-level mode prediction}      & $\argmax_k (\hat{p}_{i1}, ..., \hat{p}_{iK})$                & $\hat{Y}_i$                \Tstrut\Bstrut                                                                                                 \\ \hline
\textbf{Aggregate-level mode share prediction} & $P_k(\boldsymbol{X}_k|\hat{ \boldsymbol{\beta}}_k) = \sum_i^N \hat{p}_{ik}/N$                             & $Q_k(\boldsymbol{Z}|\hat{ \boldsymbol{\theta}}) = \sum_i^N \hat{p}_{ik}/N$       \Tstrut                                                                                 \\ \hline
\textbf{Variable importance}  & Standardized Beta coefficients                  & Variable importance, computed by using Gini index, out-of-bag error, and many others                    \Tstrut\Bstrut            
\\ \hline
\textbf{Variable effects}     & Sign and magnitude of Beta coefficients         & Partial dependence plots  
\Tstrut\Bstrut

\\ \hline
\textbf{Arc elasticity of feature $p$ for alternative $k$}           & $E_k(\boldsymbol{X}_{k,p}) = \frac{[P_k(\boldsymbol{X}_{k,-p}, \boldsymbol X_{k,p} \cdot (1+\Delta) | \hat{ \boldsymbol{\beta}}_k) - P_k(\boldsymbol{X}_{k} | \hat{ \boldsymbol{\beta}}_k)]/P_k(\boldsymbol{X}_{k} | \hat{ \boldsymbol{\beta}}_k)}{|\Delta|},$ $k \in \{1, ..., K\}$ \Tstrut &                      
$E_k(\boldsymbol{Z}_{p}) = \frac{[Q_k(\boldsymbol{Z}_{-p}, \boldsymbol Z_{p} \cdot (1+\Delta) | \hat{ \boldsymbol{\theta}}_k) - Q_k(\boldsymbol{Z} | \hat{ \boldsymbol{\theta}}_k)]/Q_k(\boldsymbol{Z} | \hat{ \boldsymbol{\theta}}_k)}{|\Delta|},$ $k \in \{1, ..., K\}$ \Tstrut
\\ \hline
\textbf{Marginal effects of feature $p$ for alternative $k$}     & $M_k(\boldsymbol{X}_{k,p}) = \frac{P_k(\boldsymbol{X}_{k,-p}, \boldsymbol{X}_{k,p} +\Delta) | \hat{ \boldsymbol{\beta}}_k) - P_k(\boldsymbol{X}_{k} | \hat{ \boldsymbol{\beta}}_k)}{|\Delta|},$ 
$k \in \{1, ..., K\}$ \Tstrut & 
$M_k(\boldsymbol{Z}_{p}) = \frac{Q_k(\boldsymbol{Z}_{-p}, \boldsymbol Z_{p} + \Delta) | \hat{ \boldsymbol{\theta}}_k) - Q_k(\boldsymbol{Z} | \hat{ \boldsymbol{\theta}}_k)}{|\Delta|},$ 
$k \in \{1, ..., K\}$                                 \Tstrut       \\ \hline
\end{tabular}
}
\label{tab:comparison}
\end{table}

\subsection{Model Development}

Logit models and machine-learning models approach the mode choice prediction problem from different perspectives. Logit models view the mode choice problem as individuals  selecting a mode from a set of travel options in order to maximize their utility. Under the random utility maximization framework, the model assumes that each mode provides a certain level of (dis)utility to a traveler, and specifies, for each mode, a utility function with two parts: A component to represent the effects of observed variables and a random error term to represent the effects of unobserved factors \citep{ben1985discrete}. For example, the utility of choosing mode $k$ under the MNL model can be defined as 
\begin{equation}
    U_{k}(\boldsymbol{X}_k|\boldsymbol{\beta}_k) = \boldsymbol{\beta}_k^T \boldsymbol{X}_{k} + \boldsymbol \varepsilon_{k},
\end{equation}
where $\boldsymbol{\beta}_k$ are the coefficients to be estimated and $\boldsymbol \varepsilon_{k}$ is the unobserved random error for choosing mode $k$. Different logit models are formed by specifying different types of error terms and different choices of coefficients on the observed variables. For instance, assuming a Gumbel distributed error term and fixed model coefficients (i.e., coefficients that are the same for all individuals) produces the MNL model \citep{ben1985discrete}. In the MNL, the probability of choosing alternative $k$ for individual $i$ is
\begin{equation}
    p_{ik} = \frac{\exp( {\boldsymbol{\beta}}_k^T \boldsymbol{X}_{ik}) }{\sum_{p=1}^K \exp ({\boldsymbol{\beta}}_p^T \boldsymbol{X}_{ik})},
\end{equation}
Given the Beta coefficient, the MNL can be associated with the likelihood function
\begin{equation}
    \boldsymbol{L}(\boldsymbol{\beta}) = \prod_{i = 1}^N \prod_{k = 1}^K \Bigg[ \frac{\exp ({\boldsymbol{\beta}}_k^T \boldsymbol{X}_{ik})}{\sum_{p=1}^K \exp ( {\boldsymbol{\beta}}_p^T \boldsymbol{X}_{ik} )} \Bigg].
\end{equation}
Maximum likelihood estimation can then be applied to obtain the ``best" utility coefficients $\hat{\boldsymbol{\beta}} = \argmax_{\boldsymbol{\beta}} \boldsymbol{L}(\boldsymbol{\beta})$. By plugging $\hat{\boldsymbol{\beta}}$ into Eqn. (2), the \textit{choice probabilities} for each mode can be obtained. More complex logit models, such as the mixed logit and nested logit, can be derived similarly from different assumptions about the coefficients and error distribution.
However, these models are more difficult to fit: They generally do not have closed-form solutions for the likelihood function and require the simulation of maximum likelihood for various parameter estimations. 
Observe also that logit models have a layer structure, which maps the input layer $\boldsymbol{X}_i$ to the output layer, $[p_{ik}, ..., p_{iK}]^T$.


Machine-learning models, by contrast, view mode choice prediction as a \textit{classification} problem: Given a set of input variables, predict which travel mode will be chosen. More precisely, the goal is to learn a target function $f$ which maps input variables $\boldsymbol{Z}$ to the output target $\boldsymbol{Y}$, $\boldsymbol{Y} \in \{1, ..., K\}$, as 
\begin{equation}
\boldsymbol{Y} = f(\boldsymbol{Z}|\boldsymbol{\theta}),
\end{equation}
where $\boldsymbol{\theta}$ represents the unknown parameter vector for parametric models like NB and the hyperparameter vector for non-parametric models such as SVM, CART, and RF. Unlike logit models that predetermine a (usually) linear model structure and make specific assumptions for parameters and error distributions, many machine-learning models are nonlinear and/or non-parametric, which allows for more flexible model structures to be directly learned from the data. In addition, compared to logit models that maximize likelihood to estimate parameters, machine-learning models often apply different optimization techniques, such as back propagation and gradient descent for NN, recursive partitioning for CART, structural risk minimization for SVM. Moreover, while logit models have a layer structure, machine-learning models have different model topologies for different models. For example, tree-based models (CART, BAG, BOOST, and RF) all have a tree structure, whereas NN has a layer structure.

Furthermore, since the outputs of logit models are individual choice probabilities, it is difficult to compare the prediction with the observed mode choices directly. Therefore, when evaluating the predictive accuracy of logit models at the individual level, a common practice in the literature is to assign an outcome probability to the alternative with the largest outcome probability, i.e.,
\begin{equation}
    \argmax_k (\hat{p}_{i1}, ...,\hat{p}_{iK}).
\end{equation}
This produces the same type of output (i.e., the travel mode choice) as the machine-learning models. Besides the prediction of individual choices, logit models and machine-learning methods are often evaluated based on their capability to reproduce the aggregate choice distribution for each mode, i.e., the market shares of each mode. For logit models, the predicted market share of mode $k$ is 
\begin{equation}
    P_k(\boldsymbol{X}_k|\hat{\boldsymbol{\beta}}_k) = \sum_i^N \hat{p}_{ik}/N,                             
\end{equation}
and, for machine-learning methods, it is usually computed by
\begin{equation}
    Q_k(\boldsymbol{Z}|\hat{\boldsymbol{\theta}}) = \sum_i^N I_k(\hat{Y}_{i})/N.
\end{equation}
However, using the proportion of predicted class labels to approximate the market share may not be ideal. Instead, similar to logit models, many machine-learning methods can directly predict class probabilities at the individual level, so in this study, we use 
\begin{equation}
    Q_k(\boldsymbol{Z}|\hat{\boldsymbol{\theta}}) = \sum_i^N \hat{p}_{ik}/N,
\end{equation}
to predict the market share for machine-learning models as well.

The calibration of the logit models is targeted at approximating aggregate market shares, as opposed to giving an absolute prediction on the individual choice \citep{ben1985discrete, hensher2005applied}. Thus, the predictive accuracy of the models may differ at the individual level and the aggregate level: Which of them should be prioritized depends on the project at hand.







\begin{figure}[!t]
    \centering
    \includegraphics[width=13cm]{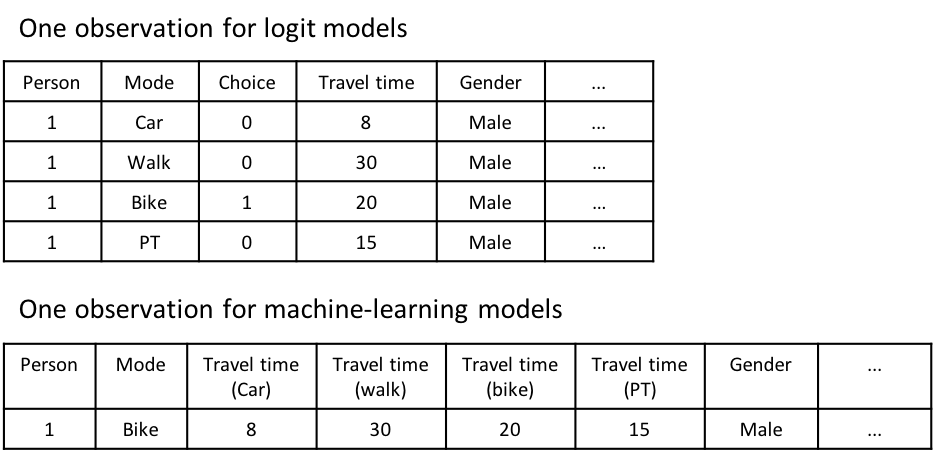}
    \caption{Data Structure for Logit Models and Machine-Learning Models.}
    \label{fig:data_input}
\end{figure}

Another important difference between the two approaches lies in the input data structures. Fitting a logit model requires the data on all available alternatives. In other words, even if the attributes of non-chosen alternatives are not observed, their values need to be imputed for the model. By contrast, machine-learning algorithms require the observed (chosen) mode only and not necessarily information on the non-chosen alternatives. Some previous studies have indeed only considered attribute values of the chosen mode, e.g., travel time of the chosen mode \citep{xie2003work, wang2018machine}, in their machine-learning models. However, a model that leaves out the attribute values of the non-chosen alternatives does not account for the fact that a given outcome is a result of the differences in attribute values across alternatives (i.e., mode competition) rather than a result of the characteristics of the chosen alternative itself. Therefore, we believe that, like logit models, the attribute values of the non-chosen alternatives should also be included (often imputed) into a machine learning model. Figure \ref{fig:data_input} shows one observation that serves as the input to logit models and machine-learning models respectively.




\subsection{Model Evaluation}



When evaluating statistical and machine-learning models, the goal is to minimize the overall prediction error, which is a sum of three terms: the bias, the variance, and the irreducible error. The bias is the error due to incorrect assumptions of the model. The variance is the error arising from the model sensitivity to the small fluctuations in the dataset used for fitting the model. The irreducible error results from the noise in the problem itself. The relationship between bias and variance is often referred to as ``bias-variance tradeoff,'' which measures the tradeoff between the goodness-of-fit and model complexity. Goodness-of-fit captures how a statistical model can capture the discrepancy between the observed values and the values expected under the model. Better fitting models tend to have more complexity, which may create overfitting issues and decrease the model predictive capabilities. On the other hand, simpler models tend to have a worse fit and a higher bias, causing the model to miss relevant relationships between input variables and outputs, which is also known as underfitting. Therefore, in order to balance the bias-variance tradeoff and obtain a model with low bias and low variance, one needs to consider multiple models at different complexity levels, and use an evaluation criterion to identify the model that minimizes the overall prediction error. The process is known as model selection. The evaluation criteria can be theoretical measures like adjusted $R^2$, AIC, $C_p$, and BIC, and/or resampling-based measures, such as cross validation and bootstrapping. Resampling-based measures are generally preferred over theoretical measures.

The selection of statistical models is usually based on theoretical measures. For example, when using logit models to predict individual mode choices, researchers usually calibrate the models on the entire dataset, examine the log-likelihood at convergence, and compare the resulting adjusted McFadden's pseudo $R^2$ \citep{mcfadden1973conditional}, AIC, and/or BIC in order to determine a best-fitting model. These three measures penalize the likelihood for including too many ``useless'' features. The adjusted McFadden's pseudo $R^2$ is most commonly reported for logit models, and a value between 0.2 to 0.3 is generally considered as indicating satisfactory model fit \citep{mcfadden1973conditional}. On the other hand, AIC and BIC are commonly used to compare models with different number of variables.


For machine learning, cross validation is usually conducted to evaluate a set of different models, with different variable selections, model types, and choices of hyper-parameters. The best model is thus identified as the one with the highest out-of-sample predictive power. A commonly-used cross validation method is the 10-fold cross validation, which applies the following procedure: 1) Randomly split the entire dataset into 10 disjoint equal-sized subsets; 2) choose one subset for validation, the rest for training; 3) train all the machine-learning models on one training set; 4) test all the trained models on the validation set and compute the corresponding predictive accuracy; 5) repeat Step 2) to 4) for 10 times, with each of the 10 subsets used exactly once as the validation data; and 6) the 10 validation results for each model are averaged to produce a mean estimate. Cross validation allows researchers to compare very different models together with the single goal of assessing their predictive accuracy. This paper compares the logit and machine-learning models using the 10-fold cross validation in order to evaluate their predictive capabilities at individual and aggregate levels.

Finally, when applying statistical models such as the logit models, researchers often take into account the underlying theoretical soundness and the behavioral realism of the model outputs to identify a final model (in addition to relying on the adjusted McFadden's pseudo $R^2$, AIC and/or BIC). In other words, even though balancing the bias-variance tradeoff is very important, in statistical modeling, a ``worse'' model may be preferred due to reasons like theoretical soundness and behavioral realism. For example, since worsening the performance of a travel mode should decrease its attractiveness, the utility coefficients of the level-of-service attributes such as wait time for transit should always have a negative sign. Therefore, when a ``better'' model produces a positive sign for wait time, a ``worse'' model with a negative sign for wait time may be preferred. On the other hand, for machine-learning models, the predictive accuracy is typically the sole criterion for deciding the best model in the past, but with the recent development of machine-learning interpretation, some researchers suggested that machine-learning models should be evaluated by both predictive accuracy and descriptive accuracy \citep{murdoch2019interpretable}.


\subsection{Model Interpretation and Application}

The interpretation of outputs of logit models is straightforward and intuitive. Like any other statistical model, researchers can quickly learn how and why a logit model works by examining the sign, magnitude, and statistical significance of the model coefficients. Researchers may also apply these outputs to conduct further behavioral analysis on individual travel behavior, such as deriving marginal effect and elasticity estimates, comparing the utility differences in various types of travel times, calculating traveler willingness-to-pay for trip time and other service attributes. All of these applications can be validated by explicit mathematical formulations and derivations, which allows modelers to clearly explain what happens ``behind the scene.''

By contrast, machine-learning models are often criticized for being ``black-box'' and lacking explanation \citep{klaiber2011random}. The lack of interpretability is believed to be a major barrier for machine learning in many real-world applications. Recently, more attention has been paid to explaining machine-learning models, with a variety of machine-learning interpretation tools being invented \citep[e.g.][]{friedman2001greedy,goldstein2015peeking,molnar2018interpretable}.
The most commonly used machine-learning interpretation tools include variable importance and partial dependence plots \citep{molnar2018interpretable}. Variable importance measures show the relative importance of each input variable in predicting the response variable. Different machine-learning models have different ways to compute variable importance. For example, for tree-based models (such as CART and RF), the mean decrease in node impurity (measured by the Gini index) is commonly used to measure the variable importance. Partial dependence plots measure the influence of a variable $\boldsymbol Z_p$ on the log-odds or probability of choosing a mode $k$ after accounting for the average effects of the other variables \citep{friedman2001greedy}. Notably, partial dependence plots may reveal causal relationships if the machine-learning model is accurate and the domain knowledge supports the underlying causal structure \citep{zhao2017causal}.



Arguably, the behavioral insights that one can extract from the logit models (such as marginal effects and elasticities) may also be obtained from machine-learning models by performing a sensitivity analysis. For example, for machine-learning models, the arc elasticity for feature $p$ of alternative $k$ can be obtained by
\begin{equation}
    E_k(\boldsymbol{Z}_{p}) = \frac{[Q_k(\boldsymbol{Z}_{-p}, \boldsymbol Z_{p} \cdot (1+\Delta) | \hat{ \boldsymbol{\theta}}) - Q_k(\boldsymbol{Z} | \hat{ \boldsymbol{\theta}})]/Q_k(\boldsymbol{Z} | \hat{ \boldsymbol{\theta}})}{|\Delta|},
\end{equation}
and the marginal effect for feature $p$ of alternative $k$ can be computed as
\begin{equation}
    M_k(\boldsymbol{Z}_{p}) = \frac{Q_k(\boldsymbol{Z}_{-p}, \boldsymbol Z_{p} + \Delta) | \hat{ \boldsymbol{\theta}}) - Q_k(\boldsymbol{Z} | \hat{ \boldsymbol{\theta}})}{|\Delta|}.
\end{equation}

In essence, all of these techniques, despite their obvious differences,  measure how the output variable responds to changes in the input features. In the context of travel mode choices, they help researchers gain a better understanding of how individual choices of travel modes are impacted by a variety of different factors such as the socio-economic and demographic characteristics of travelers and the respective trip attributes for each travel mode. In the current literature, however, the behavioral findings gained from machine-learning models are rarely compared with those obtained from logit models. Since the goals of mode choice studies are often in extracting knowledge to shed light on individual travel preferences and travel behavior instead of merely predicting their mode choice, these comparisons are necessary to have a more thorough evaluation of the adequacy of machine learning. Machine-learning models that have excellent predictive power but generate unrealistic behavioral results may not be useful in travel behavior studies.

\section{The Data for Empirical Evaluation}

\begin{figure}[!t]
    \centering
    \includegraphics[width=13cm]{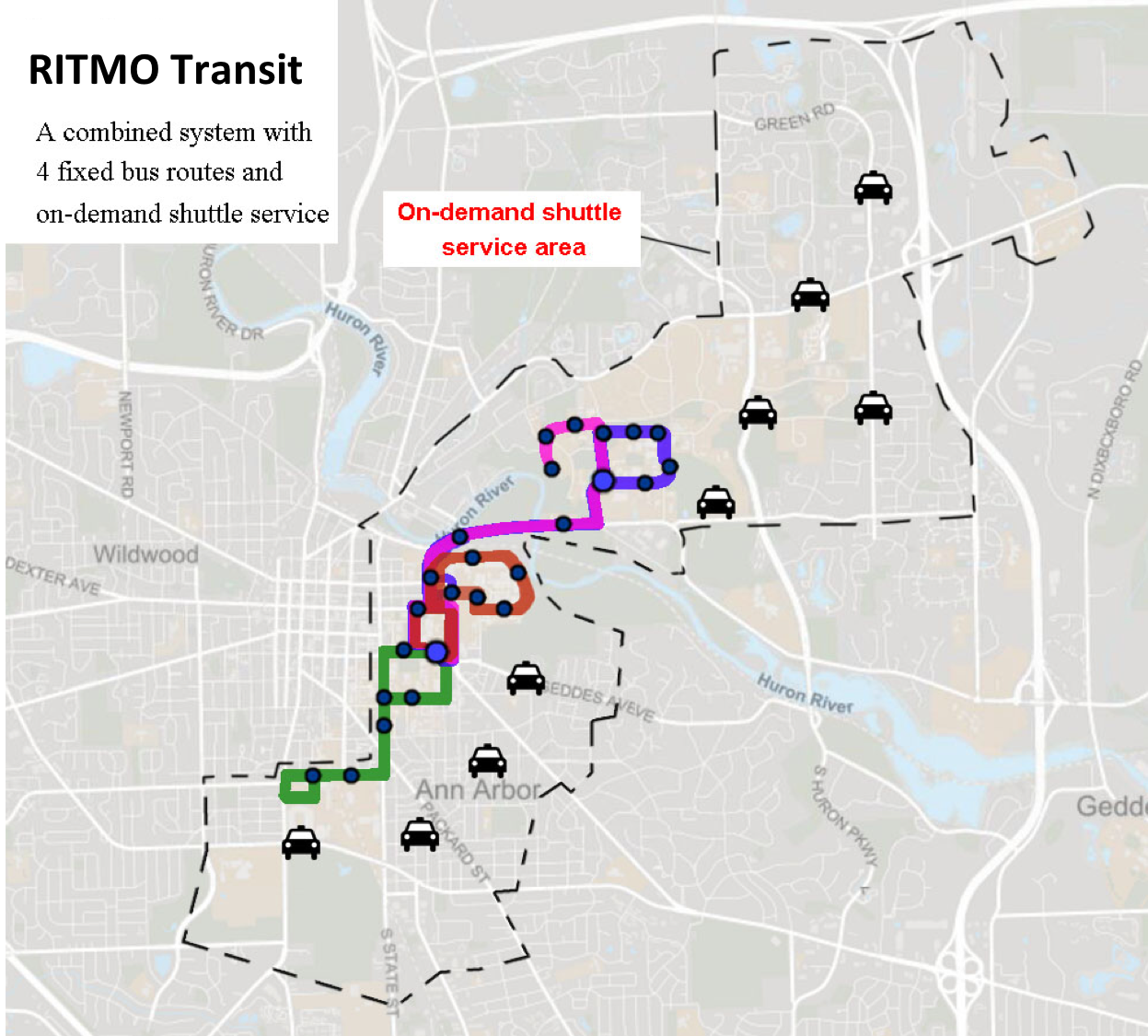}
    \caption{RITMO Transit: The New Transit System Featured with Four High-Frequency Bus Routes and On-Demand Shuttles Serving Approximately 2-Mile-Radius Area of the University of Michigan Campus.}
    \label{fig:map}
\end{figure}

The data used for empirical evaluation came from a stated-preference (SP) survey completed by the faculty, staff, and students at the University of Michigan on the Ann Arbor campus. In the survey, participants were first asked to estimate the trip attributes (e.g., travel time, cost, and wait time) for their home-to-work travel for each of the following modes: Walking, biking, driving, and taking the bus. Then, the survey asked respondents to envision a change in the transit system, i.e., the situation where a new public transit (PT) system, named RITMO Transit \citep{RitmoTransit}, fully integrating high-frequency fixed-route bus services and micro-transit services, has replaced the existing bus system (see Figure \ref{fig:map}). Text descriptions were coupled with graphical illustrations to facilitate the understanding of the new system. Each survey participant was then asked to make their commute-mode choice among \textit{Car}, \textit{Walk}, \textit{Bike}, and \textit{PT} in seven state-choice experiments, where the trip attributes for \textit{Walk}, \textit{Bike}, and \textit{Car} were the same as their self-reported values and the trip attributes for \textit{PT} were pivoted from those of driving and taking the bus. A more detailed descriptions of the survey can be found in \citet{YAN2018}.

A total of 8,141 observations collected from 1,163 individuals were kept for analysis after a data-cleaning process. The variables that enter into the analysis include the trip attributes for each travel mode, several socio-demographic variables, transportation-related residential preference variables, and current/revealed travel mode choices. The travel attributes include travel time for all modes, wait time for \textit{PT}, daily parking cost for driving, number of additional pickups for \textit{PT}, and number of transfers for \textit{PT}. The socio-economic and demographic variables include car access (car ownership for students and car per capita in the household for faculty and staff), economic status (living expenses for students and household income for faculty and staff), gender, and identity status (i.e., faculty vs staff vs student). The transportation-related residential preference variables are the importance of walkability/bikeability and transit availability when deciding where to live. Finally,  current travel mode choices are also included as state-dependence effects (i.e., the tendency for individuals to abandon or stick with their current travel mode) are verified as important predictors of mode choice by many empirical studies. Table \ref{tab:var} summarizes the descriptive statistics on these variables, including a general description of each variable, category percentages for categorical variables, and min, max, mean, and standard deviation for continuous variables.

\begin{table}[!t]
\footnotesize
\caption{Statistics for Independent Variables and Dependent Variable (Travel Mode)}
\setlength{\tabcolsep}{7pt}
\centering
\resizebox{1\textwidth}{!}{
     \begin{tabular}{ l | l | l l l l l l }
        \hline
         \textbf{Variable} & \textbf{Description} & \textbf{Category} & \textbf{\%} & \textbf{Min} & \textbf{Max} & \textbf{Mean} & \textbf{SD}  \\[0.5ex]  \hline
         \textit{Dependent Variable} &&&&&&& \\[0.5ex]
         Mode Choice  & & Car  & 14.888  &        &       &  & \\
                      &   & Walk & 28.965  &        &       &  & \\  
            &    & Bike & 20.870  &        &       &  & \\
             &   & PT   & 35.278  &        &       &  & \\
        & &&&&&& \\
        \textit{Independent Variables} &&&&&&& \\[0.5ex]
         TT\_Drive & Travel time of driving (min) &  &  & 2.000 & 40.000 & 15.210 & 6.616
         \\
         TT\_Walk & Travel time of walking (min)  &  &  & 3.000 & 120.000 & 32.300 & 23.083
         \\
         TT\_Bike & Travel time of biking (min) &  &  & 1.000 & 55.000 & 15.340 & 10.447
         \\
         TT\_PT & Travel time of using PT (min)    &  &  & 6.200 & 34.000 & 18.680 & 4.754
         \\
         Parking\_Cost & Parking cost (\$) &  &  & 0.000 & 5.000 & 0.9837 & 1.678
         \\
         Wait\_Time & Wait time for PT (min) &  &  & 3.000 & 8.000  & 5.000 & 2.070
          \\
         Transfer & Number of transfers &  &  & 0.000 & 2.000  & 0.328 & 0.646
         \\
         Rideshare & Number of additional pickups &  &  & 0.000 & 2.000  & 1.105 & 0.816
         \\
         Income& Income level
      &  &  & 1.000 & 6.000  & 1.929 & 1.342 \\
         Bike\_Walkability& Importance of bike- and walk-ability   &  &  & 1.000 & 4.000  & 3.224 & 0.954 \\
         PT\_Access& Importance of PT access  &  &  & 1.000 & 4.000  & 3.093 & 1.023  \\
         CarPerCap & Car per capita  &  &  & 0.000 & 3.000  & 0.529 & 0.476
          \\
         Female & Female or male  & Female   & 56.320   & & & & \\
                  & & Male & 43.680  & & & & \\
         Current\_Mode\_Car  & Current travel mode is Car or not & Car & 16.681 & & & & \\
         && Not Car & 83.319 & & & & \\
         Current\_Mode\_Walk  & Current travel mode is Walk or not & Walk & 40.413 & & & & \\
         && Not Walk & 59.587 & & & & \\
         Current\_Mode\_Bike  & Current travel mode is Bike or not & Bike & 8.254 & & & & \\
         && Not Bike & 91.746 & & & & \\
        Current\_Mode\_PT\footnote{Current\_Mode\_PT is not included for machine-learning models, since it can be represented by a linear combination of Current\_Mode\_Car, Current\_Mode\_Walk, and Current\_Mode\_Bike.}  & Current travel mode is PT or not & PT & 34.652 & & & & \\
         && Not PT & 65.348 & & & & \\
        [1ex]
         \hline
    \end{tabular}
    }
     \label{tab:var}
\end{table}

After extracting the data from the SP survey, we pre-processed the data and verified that all the independent variables have little multicollinearity \citep{farrar1967multicollinearity}. The existence of multicollinearity can inflate the variance and negatively impact the predictive power of the models. This study chose the variance inflation factor to determine which variables are highly correlated with other variables and found out that all variables had a variance inflation factor value of less than five, indicating that multicollinearity was not a concern.

\section{Models Examined and Their Specifications}
\label{exp.setting}

This section briefly introduces the logit and machine-learning models examined in this study. Since our dataset has a panel structure, usually a mixed logit model should be applied. However, we also fitted an MNL model as the benchmark for comparison, as previous studies generally compared machine-learning models with the MNL model only. Seven machine-learning models are examined, including simple ones like NB and CART, and more complex ones such as RF, BOOST, BAG, SVM, and NN. Most previous mode choice studies only examined a subset of these models \citep{xie2003work, omrani2013prediction, omrani2015predicting, wang2018machine, chen2017understanding}.

\subsection{Logit Models}

We have already introduced the MNL model formulation in detail in Subsection 3.1, so only the mixed logit model is presented here.

The mixed logit model is an extension of the MNL model, which addresses some of the MNL limitations (such as relaxing the IIA property assumption) and is more suitable for modeling panel choice datasets in which the observations are correlated (i.e., each individual is making multiple choices) \citep{mcfadden2000mixed}. A mixed logit model specification usually treats the coefficients in the utility function as varying across individuals but being constant over choice situations for each person \citep{train2009discrete}. The utility function from alternative $k$ in choice occasion $t$ by individual $i$ is
\begin{equation}
    U_{ikt} = \boldsymbol{\beta}_{ik}^T \boldsymbol{X}_{ikt} + \varepsilon_{ikt},
\end{equation}
where $\varepsilon_{ikt}$ is the independent and identically distributed random error across people, alternatives, and time. Hence, conditioned on $\boldsymbol{\beta}$, the probability of an individual making a sequence of choices (i.e., $\boldsymbol{j} = \{j_1, j_2, ..., j_\tau\}$) is
\begin{equation}
    \boldsymbol{L}_{i\boldsymbol{j}}(\boldsymbol{\beta}) = \prod_{t = 1}^\tau \Bigg[ \frac{ \exp(\boldsymbol{\beta}_{ij_t}^T \boldsymbol{X}_{i j_{t} t}) }{ \sum_k \exp(\boldsymbol{\beta}_{ik}^T \boldsymbol{X}_{ikt})} \Bigg].
\end{equation}
Because the $\varepsilon_{ikt}$'s are independent over the choice sequence, the corresponding unconditional probability is
\begin{equation}
    p_{ik\boldsymbol{j}} = \int \boldsymbol{L}_{i\boldsymbol{j}}(\boldsymbol{\beta}) g(\boldsymbol{\beta}) d\boldsymbol{\beta},
\end{equation}
where $g(\boldsymbol{\beta})$ is the probability density function of $\boldsymbol{\beta}$. This integral does not have an analytical solution, so it can only be estimated using simulated maximum likelihood \citep[e.g.][]{train2009discrete}.

In this study, the MNL models can be summarized as follows: 1) The utility function of \textit{Car} includes mode-specific parameters for TT\_Drive, Parking\_Cost, Income, CarPerCap, and Current\_Mode\_Car; 2) the utility function of \textit{Walk} includes mode-specific parameters for TT\_Walk, Female (sharing the same parameter with \textit{Bike}), Bike\_Walkability (sharing the same parameter with \textit{Bike}), and Current\_Mode\_Walk; 3) the utility function of \textit{Bike} includes mode-specific parameters for TT\_Bike, Female (sharing the same parameter with \textit{Walk}), Bike\_Walkability (sharing the same parameter with \textit{Walk}), and Current\_Mode\_Bike; and 4) the utility function of \textit{PT} includes mode-specific parameters for TT\_PT, Wait\_Time, Rideshare, Transfer, PT\_Access, and Current\_Mode\_PT. 
We also specify three alternative-specific constants for \textit{Walk}, \textit{Bike}, and \textit{PT}, respectively.

The mixed logit model has the same model specification. Moreover, in order to accommodate individual preference heterogeneity (i.e., taste variations among different individuals), coefficients on the selected level-of-service variables (i.e., TT\_PT and Parking\_Cost) are also specified as random parameters. The alternative-specific constant for \textit{PT} is also assumed as a random parameter. These random parameters are all assessed with a normal distribution. We use 1,000 Halton draws to perform the numerical integration. Both the MNL and mixed logit models are estimated using the NLOGIT software.

\subsection{Machine-Learning Models}
\label{subsec4}

\subsubsection{Naive Bayes}

The NB model is a simple machine-learning classifier. The model is constructed using Bayes' Theorem with the naive assumption that all the features are independent \citep{mccallum1998comparison}. NB models are useful because they are faster and easier to construct as compared to other complicated models. As a result, NB models work well as a baseline classifier for large datasets. In some cases, NB even outperforms more complicated models \citep{zhang2004optimality}. A limitation of the NB model is that, in real world situations, it is very unlikely for all the predictors to be completely independent from each other. Thus, the NB model is very sensitive when there are highly correlated predictors in the model. In this study, the NB model is constructed through the R package \textit{e1071} \citep{e1071}.

\subsubsection{Tree-based Models}

The CART model builds classification or regression trees to predict either a classification or a continuous dependent variable. In this paper, the CART model creates classification trees where each internal node of the tree recursively partitions the data based on the value of a single predictor. Leaf nodes represent the category (i.e., \textit{Car}, \textit{Bike}, \textit{PT}, and \textit{Walk}) predicted for that individual \citep{breiman2017classification}. The decision tree is sensitive to noise and susceptible to overfit \citep{last2002improving,quinlan2014c4}. To control its complexity, it can be pruned. This study prunes the tree until the number of terminal nodes is 6. The CART model is obtained through the R package \textit{tree} \citep{tree}.

To address the overfitting issues of CART models, the tree-based ensemble techniques were proposed to form more robust, stable, and accurate models than a single decision tree \citep{breiman1996bagging, friedman2001elements}. One of these ensemble methods is BOOST. For a $K$-class problem,  BOOST creates a sequence of decision trees, where each successive tree seeks to improve the incorrect classifications of the previous trees. Predictions in BOOST are based on a weighted voting among all the boosting trees. Although  BOOST usually has a higher predictive accuracy than CART, it is more difficult to interpret. Another drawback is that BOOST is prone to overfitting when too many trees are used. This study applies the gradient boosting machine technique to create the BOOST model \citep{friedman2001greedy}. 400 trees are used, with shrinkage parameter set to 0.14 and the interaction depth to 10. The minimum number of observations in the trees terminal nodes is 10. The BOOST model is created with the R package \textit{gbm} \citep{gbm}.

Another well-known ensemble method is BAG, which trains multiple trees in parallel by bootstrapping data (i.e., sampling with replacement) \citep{breiman1996bagging}. The BAG model uses all the independent variables to train the trees. For a $K$-class problem, after all the trees are trained, the BAG model makes the mode choice prediction by determining the majority votes among all the decision trees. By using bootstrapping, the BAG model is able to reduce the variance and overfitting problems of a single decision tree model. One potential drawback with the BAG model is that it assumes that all the features are independent. If the features are correlated, the variance would not be reduced with BAG. In this study, 400 classification trees are bagged, with each tree grown without pruning.

The RF model is also an ensemble method. Like BAG, RF trains multiple trees using bootstrapping \citep{ho1998random}. However, RF only uses a random subset of all the independent variables to train the classification trees. More precisely, the trees in RF use all the independent variables, but every node in each tree only uses a random subset of them \citep{breiman2001random}. By doing so, RF reduces variance between correlated trees and negates the drawback that BAG models may have with correlated variables. Similar to BAG, RF makes mode choice predictions by determining the majority voting among all the classification trees. Like other non-parametric models, RF is difficult to interpret. In this study, 500 trees are used and 12 randomly selected variables are considered for each split at the trees' nodes. The R package used for producing the BAG and RF models is \textit{randomForest} \citep{RF}.

\subsubsection{Support Vector Machine}

The SVM model is a binary classifier which, given labeled training data, finds the hyper-plane maximizing the margin between two classes. This hyperplane is a linear or nonlinear (depending on the kernel) decision boundary that separates the two classes. Since a mode choice model typically involves multi-class classification, the one-against-one approach is used \citep{hsu2002comparison}. Specifically, for a $K$-class problem, $K(K-1)/2$ binary classifiers are trained to differentiate all possible pairs of $K$ classes. The class receiving the most votes among all the binary classifiers is selected for prediction. SVM usually performs well with both nonlinear and linear boundaries depending on the specified kernel. However, the SVM model can be very sensitive to overfitting especially for nonlinear kernels \citep{cawley2010over}. In this study, a SVM with a radial basis kernel is used. The cost constraint violation is set to 8, and the gamma parameter for the kernel is set to 0.15. The SVM model is produced with the R package \textit{e1071} \citep{e1071}.

\subsubsection{Neural Network}

A basic NN model has three layers of units/nodes where each node can either be turned active (on) or inactive (off), and each node connection between layers has a weight. The data is fed into the model at the input layer, goes through the weighted connections to the hidden layer, and lastly ends up at a node in the output layer which contains $K$ units for an $K$-class problem. The hidden layer allows the NN to model nonlinear relationships between variables. Although NN has shown promising results in modeling travel mode choice in some studies \citep{omrani2015predicting}, NN models tend to be overfitting, and are difficult to interpret. In this paper, a NN with a single hidden layer of 18 units is used. The connection weights are trained by back propagation with a weight decay constant of 0.4. The R package \textit{nnet} is used to create our NN model \citep{stats}.

\section{Comparison of Empirical Results}

This section presents the empirical results of this study. Specifically, it compares the predictive accuracy of the logit models with that of the machine-learning algorithms. In addition, it compares the behavioral findings of two machine-learning models (i.e., RF and NN) and two logit models (i.e., MNL and mixed logit). 


\subsection{Predictive Accuracy}

This study applied the 10-fold cross validation approach. As discussed above, cross validation requires splitting the sample data into training sample sets and validation sample sets. One open issue is how to partition the sample dataset when it is a panel dataset (i.e., individuals with multiple observations). One approach is to treat all observations as independent choices and randomly divide these observations. The other is to subset by individuals, each with their full set of observations. This study follows the first approach, which is commonly applied by previous studies \citep{xie2003work, hagenauer2017comparative, wang2018machine}. 

As discussed in Subsection 3.1, the predictive power of the models may differ at the individual level (predicting the mode choice for each observation) and at the aggregate level (predicting the market shares for each travel mode). The calibration of logit models focuses on reproducing market shares whereas the development of machine-learning classifiers aims at predicting individual choices. This study compares both the mean individual-level predictive accuracy and the mean aggregate-level predictive accuracy.

\subsubsection{Individual-Level Predictive Accuracy}

The cross validation results for individual-level predictive accuracy is shown in Table \ref{tab:accuracy_ind_out}. The best-performing model is RF, with a mean predictive accuracy equal to 0.856. However, the accuracy of the MNL and the mixed logit model is only 0.647 and 0.631 respectively, which is much lower than the RF model.


The predictive accuracy of each model by travel mode is also presented in Table \ref{tab:accuracy_ind_out}. All models predict {\em Walk} most accurately. All machine-learning models have a mean predictive accuracy value between 0.795 and 0.928, whereas the MNL model has an accuracy of 0.859 and the mixed logit model 0.797. Both logit models and three ensemble machine-learning models (i.e., BOOST, BAG, and RF) predict modes {\em PT} and {\em Bike} relatively better than mode {\em Car}. One possible explanation is that {\em Car}, with a market share of 14.888\%, has fewer observations compared to other modes. The notorious class imbalance problem may cause machine-learning classifiers to have more difficulties in predicting the class with fewer observations.

Finally, it is somewhat surprising that the mixed logit model, a model that accounts for individual heterogeneity and has significantly better model fit (adjusted McFadden's pseudo $R^2$ is 0.536) than the MNL model (adjusted McFadden's pseudo $R^2$ is 0.365), underperformed the MNL model in terms of the out-of-sample predictive power. This finding is nonetheless consistent with the findings of \citet{cherchi2010validation}. It suggests that the mixed logit model may have overfitted the data with the introduction of random parameters, and such overfitting resulted in greater out-of-sample prediction error. 




\begin{table}[!t]
\setlength{\tabcolsep}{6pt}
\centering
\begin{tabular}{ccccccccccc}
\hline
\multirow{2}{*}{Model} & \multicolumn{2}{c}{All} & \multicolumn{2}{c}{Car} & \multicolumn{2}{c}{Walk} & \multicolumn{2}{c}{Bike} & \multicolumn{2}{c}{PT} \\ \cline{2-11} 
                       & Mean       & SD         & Mean       & SD         & Mean        & SD         & Mean        & SD         & Mean       & SD        \\ \hline
MNL                    & 0.647      & 0.016      & 0.440      & 0.044      & 0.859       & 0.018      & 0.414       & 0.033      & 0.698      & 0.029     \\
Mixed logit            & 0.631      & 0.008      & 0.513      & 0.031      & 0.797       & 0.014      & 0.413       & 0.038      & 0.673      & 0.027     \\
NB                     & 0.584      & 0.018      & 0.558      & 0.035      & 0.864       & 0.013      & 0.372       & 0.041      & 0.490      & 0.042     \\
CART                   & 0.593      & 0.014      & 0.428      & 0.032      & 0.795       & 0.022      & 0.329       & 0.038      & 0.653      & 0.026     \\
BOOST                  & 0.850      & 0.007      & 0.790      & 0.035      & 0.913       & 0.012      & 0.848       & 0.023      & 0.825      & 0.028     \\
BAG                    & 0.854      & 0.013      & 0.791      & 0.017      & 0.926       & 0.016      & 0.861       & 0.028      & 0.818      & 0.029     \\
RF                     & 0.856      & 0.012      & 0.797      & 0.022      & 0.928       & 0.016      & 0.859       & 0.021      & 0.820      & 0.027     \\
SVM                    & 0.772      & 0.012      & 0.701      & 0.027      & 0.878       & 0.026      & 0.681       & 0.033      & 0.770      & 0.026     \\
NN                     & 0.646      & 0.016      & 0.434      & 0.045      & 0.853       & 0.025      & 0.451       & 0.051      & 0.679      & 0.024     \\ \hline
\end{tabular}

\caption{Mean Out-of-Sample Accuracy of Logit and Machine-Learning Models (Individual Level)}
\label{tab:accuracy_ind_out}
\end{table}

\subsubsection{Aggregate-Level Predictive Accuracy}

We now turn to aggregate-level predictive accuracy. To quantify the sum of the absolute differences between the market share predictions and the real market shares from the validation data, we use the L1-norm, also known as the least absolute deviations. Taking machine-learning models as an example, let $Q_k^*$ and $\hat{Q}_k = Q_k(\boldsymbol{Z}|\hat{\boldsymbol \theta})$ represent the true (observed) and predicted market shares for mode $k$. The L1-norm thus is defined as
\begin{equation}
    \sum_{k = 1}^4 |Q_k^* - \hat{Q}_k|.
\end{equation}

\noindent
The predictive accuracy results of the logit and machine-learning models at the aggregate level are depicted in Table \ref{tab:accuracy_agg_ave}. The results show that RF outperforms all the other models, with a prediction error of 0.0248 and a standard deviation of 0.0128. Notably, even though logit models are expected to have good performance for market share predictions, RF has lower error compared to MNL (0.0399) and mixed logit (0.0593). Again, the MNL model resulted in a higher aggregate-level predictive accuracy than the mixed logit model.

\begin{table}[!t]
\caption{Mean L1-Norm Error for Mode Share Prediction}
\setlength{\tabcolsep}{6pt}
\centering
\begin{tabular}{ccc}
\hline
Model       & Mean   & SD     \\ \hline
MNL         & 0.0399 & 0.0207 \\
Mixed logit & 0.0593 & 0.0268 \\
NB          & 0.2771 & 0.0363 \\
CART        & 0.0463 & 0.0280 \\
BOOST       & 0.0291 & 0.0151 \\
BAG         & 0.0253 & 0.0130 \\
RF          & 0.0248 & 0.0128 \\
SVM         & 0.0362 & 0.0218 \\
NN          & 0.0493 & 0.0196 \\ \hline
\end{tabular}
\label{tab:accuracy_agg_ave}
\end{table}

In summary, the results show that RF is the best model among all models evaluated and that logit models only outperform a minority of the machine learning models.

\subsection{Model Interpretation}

Recent advances in machine learning make models interpretable through techniques such as variable importance and partial dependence plots. Machine-learning results can be readily applied to compute behavioral outputs such as marginal effects and arc elasticities. However, other behavioral outputs such as the value of time, willingness-to-pay, and consumer welfare measures are hard to obtain from machine-learning models, because they are grounded on the random utility modeling framework and an assumption that individual utility can be kept constant when attributes of a product substitutes each other (e.g., paying a certain amount of money to reduce a unit of time). Machine-learning models lacks the behavioral foundation required to obtain these measures. 

This section interprets the results of two logit models (MNL and mixed logit) and two machine-learning models (RF and NN\footnote{The reasons for choosing these two machine-learning models are: 1) RF is the best-performing model among the seven machine-learning classifiers; and 2) NN is one of the most popular machine-learning classifier used for travel mode choice modeling.}). For the logit models, we interpret the coefficient estimates and calculate some behavioral measures including marginal effects and arc elasticities. In the meantime, we conduct comparable behavioral analysis on the RF and NN models by applying variable importance and partial dependence plots and by performing a sensitivity analysis. 

It should be noted that the behavioral analysis conducted here is far from exhaustive, as mode choice model applications often go beyond what is covered here. In particular, recent advances in mode choice modeling, such as the development of mixed logit and latent class models, are mainly concerned about deriving insights on individual preference heterogeneity. In a separate paper \citep{zhao2019modeling}, we showed that machine learning algorithms can automatically capture individual heterogeneity and that individual conditional expectation plots can help visualize such results. 




\subsubsection{Variable Importance and Effects}

Generally speaking, for traditional statistical models, standardized Beta coefficients can represent the strength of the effect of each independent variable on the mode choice, and the variable with the largest standardized coefficient has the strongest influence. However, the utility of choosing a travel mode is a latent variable and thus unobservable, so it is not obvious how to standardize a latent variable in order to estimate the standardized Beta coefficients. If one is only interested in the rank order of the magnitude of the effects of the independent variables on the utility, the $X$-standardization is enough and easy-to-implement, by standardizing the input variables only when conducting estimation \citep{menard2004six}. To be specific, the $X$-standardized Beta coefficients of logit models represent the weights and direction of the input variables to show the magnitude and direction of their effects.

The outputs for the MNL and mixed logit are presented in Table \ref{tab:logit_results}. The adjusted McFadden's pseudo $R^2$ for MNL and mixed logit are 0.365 and 0.536, which indicates satisfactory model fit. All the coefficient estimates are consistent with theoretical predictions. All the level-of-service variables carry an intuitive negative sign, and all of them are statistically significant. For both logit models, individual socio-demographic characteristics are associated with their travel mode choices. Unsurprisingly, higher-income travelers with better car access are more likely to drive than using alternative modes. Females are less likely to choose {\em Walk} and {\em Bike} than males. The model also shows that individual residential preferences and current travel mode choices are associated with their travel mode choices of {\em Car}, {\em Walk}, and {\em Bike}. However, people tend to have weak attachment to {\em PT} as shown by the small and insignificant Beta coefficient. Individuals who value walking, biking, and transit access when choosing where to live are more likely to use these modes. The model shows that travelers tend to stick to their current mode even when a new travel option is offered. Furthermore, for the mixed logit model, the random parameter standard deviations are also statistically significant. 

We also presented $X$-standardized Beta coefficients for the MNL and mixed logit models, allowing researchers to assess the relative importance of the independent variables, i.e., a coefficient of larger magnitude indicate a greater impact of the corresponding independent variable on the choice outcome \citep{menard2004six}. For both models, the results show that the most important variable in predicting the mode choice is TT\_Bike, followed by the travel time variables for the other three modes, several revealed-preference (RP) variables (i.e., current travel modes), and some level-of-service attributes. These results are reasonable and generally consistent with findings in the existing literature.



\begin{table}[!t]
\caption{Outputs of the MNL and Mixed Logit Models.}
\centering
\resizebox{1\textwidth}{!}{
\begin{tabular}{llcccccc}
\hline
\textbf{Variable}                             & \textbf{Alternatives} & \multicolumn{3}{c}{\textbf{MNL}}                      & \multicolumn{3}{c}{\textbf{Mixed logit}}               \\ \cline{3-8} 
                                              &                       & \multicolumn{2}{c}{Unstandardized} & $X$-standardized & \multicolumn{2}{c}{Unstandardized} & $X$-standardized  \\
                                              &                       & \multicolumn{2}{c}{coefficients}   & coefficients     & \multicolumn{2}{c}{coefficients}   & coefficients      \\ \cline{3-8} 
                                              &                       & $\beta$            & S.E.          & $\beta$Std$X$    & $\beta$            & S.E.          & $\beta$Std$X$     \\ \hline
\textbf{Constants}                            &                       &                    &               &                  &                    &               &                   \\
Walk                                          & Walk                  & 2.792**            & 0.194         & /                & 2.098**            & 0.286         & /                 \\
Bike                                          & Bike                  & 1.678**            & 0.181         & /                & 0.837**            & 0.273         & /                 \\
PT                                            & PT                    & 3.224**            & 0.190         & /                & 4.510**            & 0.560         & /                 \\
                                              &                       &                    &               &                  &                    &               &                   \\
\textbf{Level-of-service variables}           &                       &                    &               &                  &                    &               &                   \\
TT\_Drive                                     & Car                   & $-0.075$**           & 0.005         & $-1.138$**         & $-0.110$**           & 0.009         & $-2.052$**          \\
TT\_Walk                                      & Walk                  & $-0.146$**           & 0.004         & $-2.203$**         & $-0.168$**           & 0.005         & $-3.158$**          \\
TT\_Bike                                      & Bike                  & $-0.162$**           & 0.006         & $-2.451$**         & $-0.205$**           & 0.008         & $-4.166$**          \\
TT\_PT                                        & PT                    & $-0.104$**           & 0.009         & $-1.563$**         & $-0.170$**           & 0.034         & $-2.931$**          \\
Wait\_Time                                    & PT                    & $-0.156$**           & 0.018         & $-0.323$**         & $-0.470$**           & 0.046         & $-1.005$**          \\
Parking\_Cost                                 & Car                   & $-0.148$**           & 0.027         & $-0.248$**         & $-0.440$**           & 0.093         & $-2.226$**          \\
Rideshare                                     & PT                    & $-0.433$**           & 0.042         & $-0.354$**         & $-1.221$**           & 0.103         & $-1.064$**          \\
Transfer                                      & PT                    & $-0.570$**           & 0.047         & $-0.368$**         & $-1.886$**           & 0.174         & $-1.298$**          \\
                                              &                       &                    &               &                  &                    &               &                   \\
\textbf{Socio-demographic variables}          &                       &                    &               &                  &                    &               &                   \\
Income                                        & Car                   & 0.075*             & 0.030         & 0.101*           & 0.026              & 0.059         & 0.318             \\
CarPerCap                                     & Car                   & 0.560**            & 0.083         & 0.267**          & 0.505**            & 0.123         & 0.357*            \\
Female                                        & Walk, Bike            & $-0.175$**           & 0.061         & $-0.087$**         & $-0.289$**           & 0.111         & $-0.043$           \\
                                              &                       &                    &               &                  &                    &               &                   \\
\textbf{Residential preference variables}     &                       &                    &               &                  &                    &               &                   \\
Bike\_Walkability                             & Walk, Bike            & 0.069*             & 0.033         & 0.066*           & 0.320**            & 0.059         & 0.398**           \\
PT\_Access                                    & PT                    & 0.113**            & 0.031         & 0.115**          & 0.203              & 0.143         & 0.115             \\
                                              &                       &                    &               &                  &                    &               &                   \\
\textbf{Current travel mode}                  &                       &                    &               &                  &                    &               &                   \\
Current\_Mode\_Car                            & Car                   & 1.366**            & 0.094         & 0.509**          & 1.418**            & 0.154         & 1.797**           \\
Current\_Mode\_Walk                           & Walk                  & 1.289**            & 0.077         & 0.633**          & 1.066**            & 0.088         & 0.478**           \\
Current\_Mode\_Bike                           & Bike                  & 2.899**            & 0.120         & 0.798**          & 3.628**            & 0.223         & 1.031**           \\
Current\_Mode\_PT                             & PT                    & 0.093              & 0.073         & 0.044            & 2.837**            & 0.362         & 1.313**           \\
                                              &                       &                    &               &                  &                    &               &                   \\
\textbf{Random parameter standard deviations} &                       &                    &               &                  &                    &               &                   \\
PT (Constant)                                         & PT                    &                    &               &                  & 3.766**            & 0.209         & 4.139**           \\
TT\_PT                                        & PT                    &                    &               &                  & 0.089**            & 0.014         & 4.664**           \\
Parking\_Cost                                 & Car                   &                    &               &                  & 0.907**            & 0.088         & 5.999**           \\
                                              &                       &                    &               &                  &                    &               &                   \\ \hline
Sample size                                   &                       & 1163               &               &                  & 1163               &               &                   \\
Log likelihood at constant                    &                       & $-11285.82$          &               &                  & $-11285.82$          &               &                   \\
Log likelihood at convergence                 &                       & $-7160.97$           &               &                  & $-5234.94$           &               &                   \\
Adjusted McFadden's pseudo $R^2$              &                       & 0.365              &               &                  & 0.536              &               &                   \\ \hline
\multicolumn{8}{l}{Note: * significant at the 5\% level, ** significant at the 1\% level.}                                                                                            
\end{tabular}

}
\label{tab:logit_results}
\end{table}

To extract similar interpretations of logit models, this study applies widely-used tools including variable importance measures and partial dependence plots to interpret the RF model and compare the behavioral findings obtained from the RF with those from the MNL models. Like the $X$-standardized Beta coefficients in a logit model, a variable importance measure can be used to indicate the impact of an input variable on predicting the response variable for machine-learning models. Unlike $X$-standardized Beta coefficients that can show the direction of association between the input variable and the outcome variable with a positive or negative sign, however, variable importance measures provide no such information and we need to use additional machine-learning interpretation tools, such as partial dependence plots, to extract these insights. This study uses the Gini index to measure variable importance for RF. For NN, the variable importance is computed using the method proposed by \citet{gevrey2003review}, which applies combinations of the absolute values of the weights.

Table \ref{tab:VarImp} shows the ranking of variable importance for RF, NN, MNL, and mixed logit. Note that Current\_Mode\_PT is not included for RF and NN.
The ranking of the input features in RF is generally consistent with that of the two logit models, but NN shows very different variable importance results compared to RF, MNL, and mixed logit in many cases. For the RF model and the two logit models, the travel times of walking, driving, biking, and transit have very high influence on their stated mode choice; on the other hand, some differences do exist: For example, PT\_Access, Bike\_Walkability, Income, and CarPerCap are more important for RF compared to MNL and mixed logit.


\begin{table}[!t]
\caption{Ranking of Variable Importance for RF, NN, MNL, and Mixed Logit}
\begin{tabular}{c|cccc}
\hline
\textbf{Variable}   & \textbf{RF} & \textbf{NN} & \textbf{MNL} & \textbf{Mixed logit} \\ \hline
TT\_Walk            & 1           & 16          & 2            & 2                    \\
TT\_Drive           & 2           & 14          & 4            & 5                    \\
TT\_Bike            & 3           & 13          & 1            & 1                    \\
TT\_PT              & 4           & 11          & 3            & 3                    \\
Current\_Mode\_Bike & 5           & 2           & 5            & 10                   \\
PT\_Access          & 6           & 8           & 13           & 16                   \\
Bike\_Walkability   & 7           & 6           & 16           & 13                   \\
Income              & 8           & 10          & 14           & 15                   \\
CarPerCap           & 9           & 7           & 11           & 14                   \\
Current\_Mode\_Walk & 10          & 1           & 6            & 12                   \\
Rideshare           & 11          & 9           & 9            & 9                    \\
Transfer            & 12          & 5           & 8            & 8                    \\
Wait\_Time          & 13          & 15          & 10           & 11                   \\
Female              & 14          & 3           & 15           & 17                   \\
Parking\_Cost       & 15          & 12          & 12           & 4                    \\
Current\_Mode\_Car  & 16          & 4           & 7            & 6                    \\
Current\_Mode\_PT   & /           & /           & 17           & 7                    \\ \hline
\end{tabular}
\label{tab:VarImp}
\end{table}

\begin{figure}[!t]
    \centering
    \begin{subfigure}[b]{0.48\textwidth}
        \centering
        \includegraphics[height=2.8in]{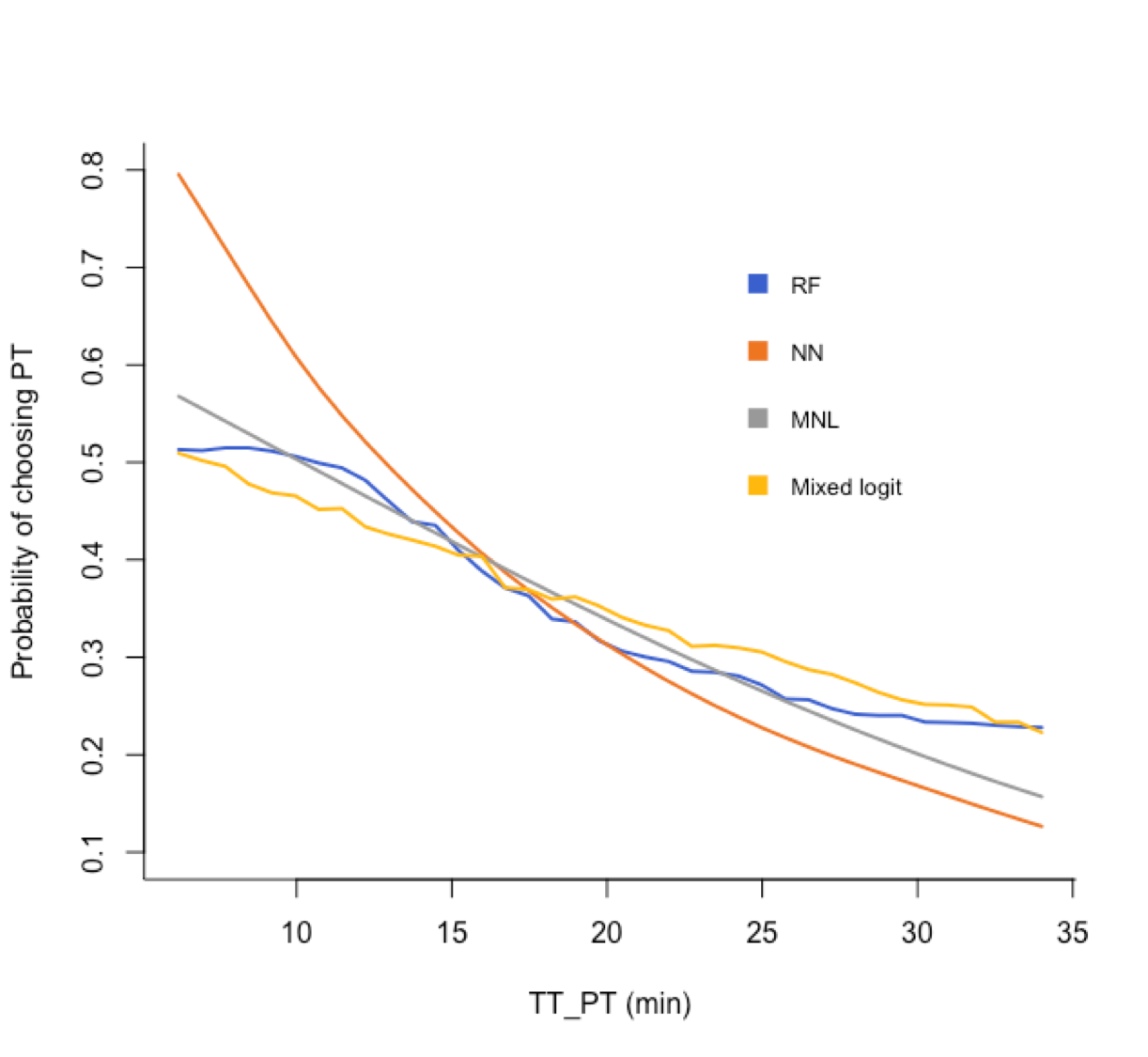}
        \caption{Partial dependence on TT\_PT}
        \label{fig:pdptt_pt}
    \end{subfigure}%
    ~
    \begin{subfigure}[b]{0.48\textwidth}
        \centering
        \includegraphics[height=2.8in]{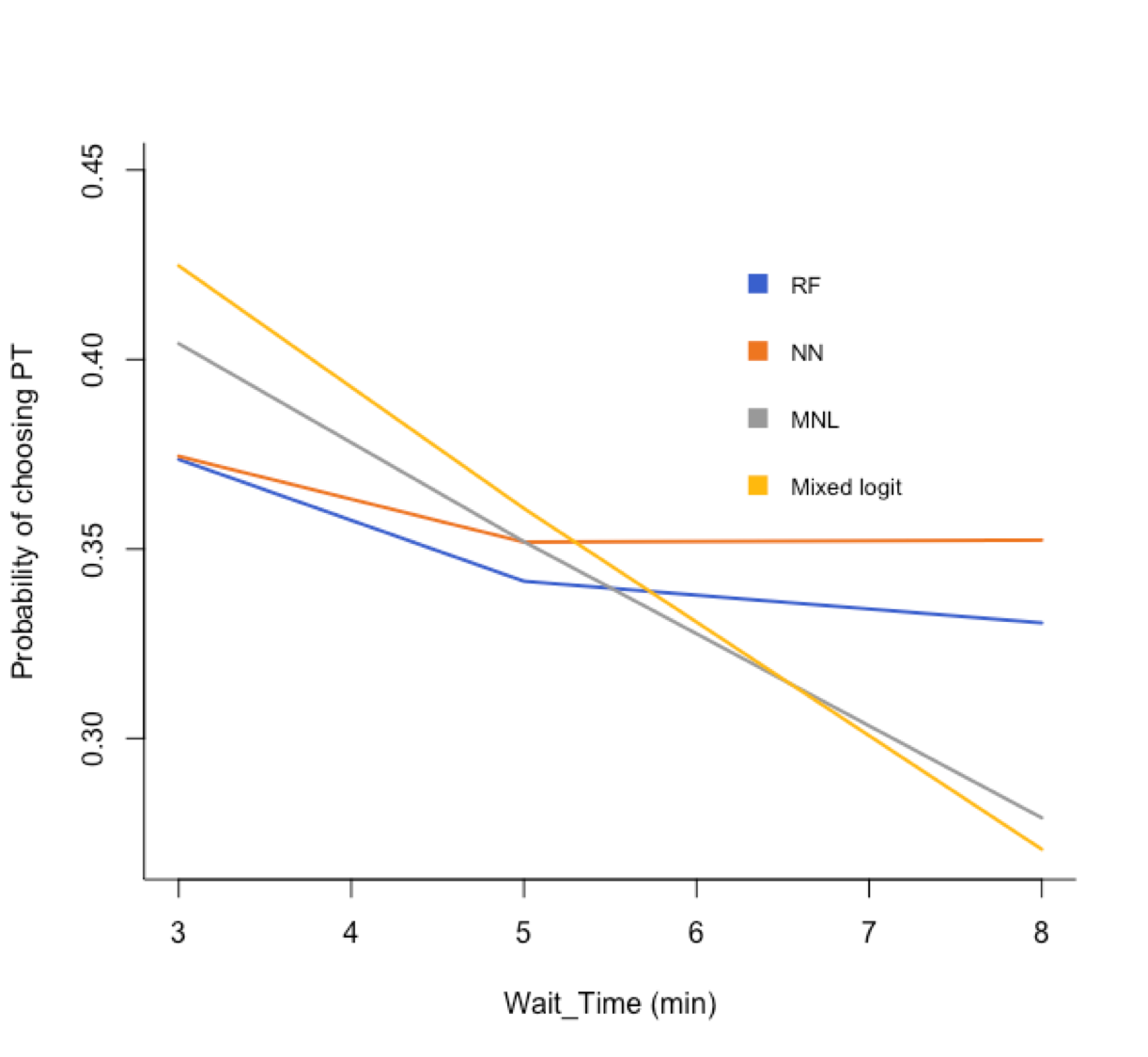}
        \caption{Partial dependence on Wait\_Time}
        \label{fig:pdpWaittime}
    \end{subfigure}%
    
    \begin{subfigure}[b]{0.48\textwidth}
        \centering
        \includegraphics[height=2.8in]{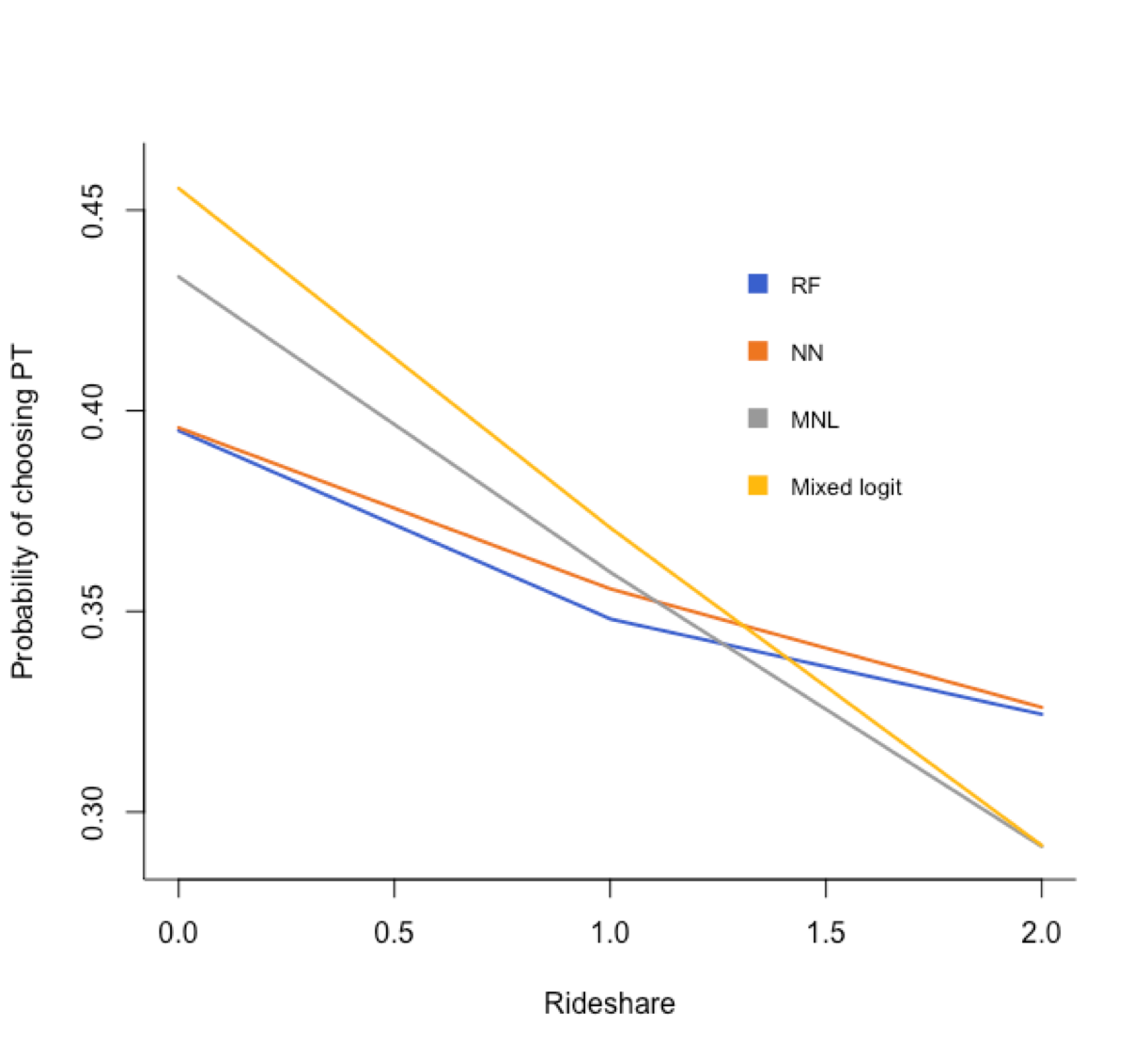}
        \caption{Partial dependence on Rideshare}
        \label{fig:pdprideshare}
    \end{subfigure}
    ~
    \begin{subfigure}[b]{0.48\textwidth}
        \centering
        \includegraphics[height=2.8in]{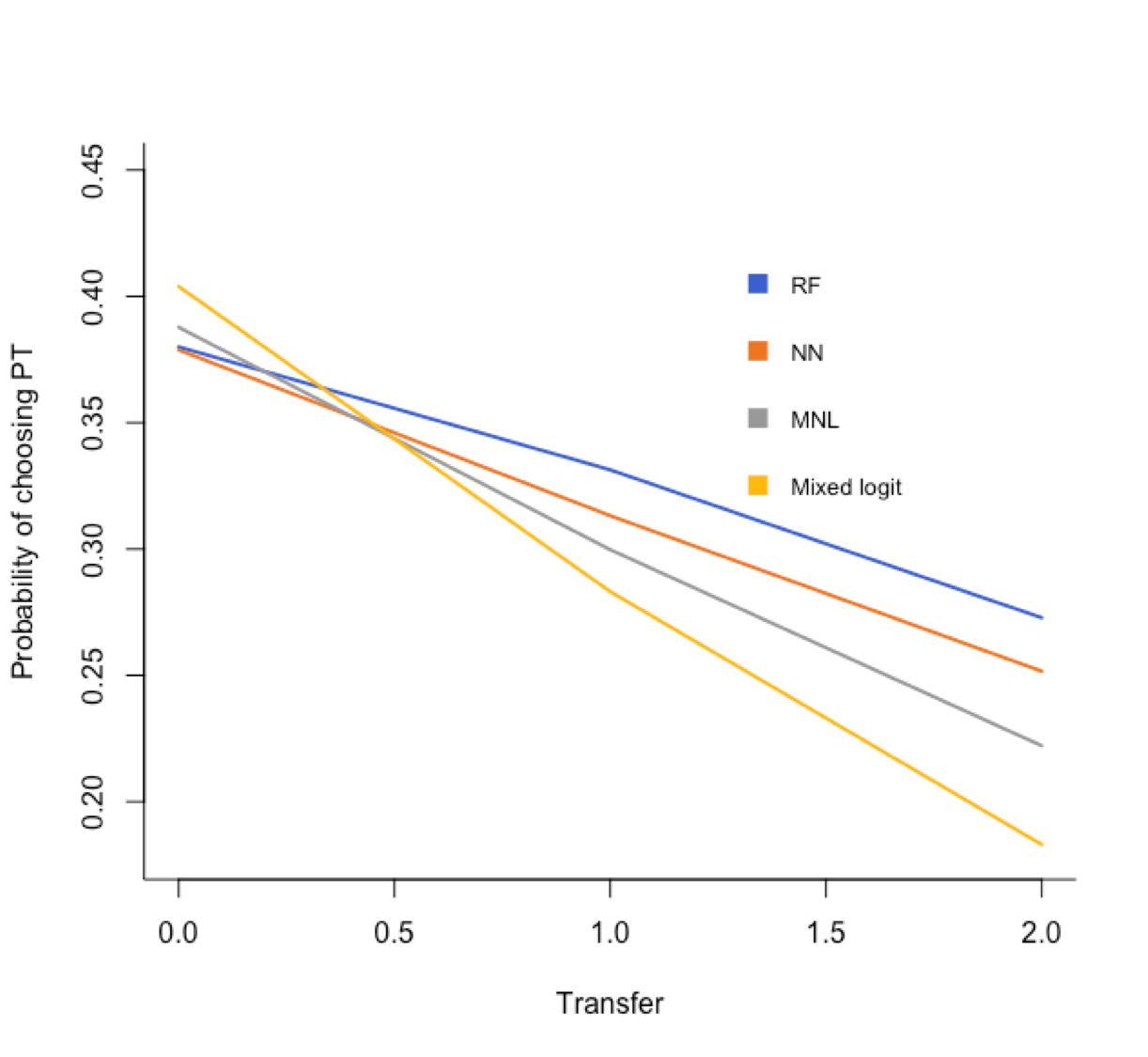}
        \caption{Partial dependence on Transfer}
        \label{fig:pdptransfer}
    \end{subfigure}
    \caption{Partial dependence plots of variables for choosing \textit{PT} as the travel mode}
    \label{fig:PDP}
\end{figure}

Partial dependence plots are another important tool that helps interpret machine-learning models. Figure \ref{fig:PDP} presents how the probability of choosing \textit{PT} changes as the value of the selected variable changes for RF and MNL. The shape of the curves sheds light on the direction and magnitude of the changes, which is similar to the Beta coefficients (without standardization) estimated from the MNL model. However, the Beta coefficients in logit models affect the utility of mode $k$ (see Eqns. (1) and (11)) rather than the probability of choosing mode $k$ (see Eqns. (2) and (13)). Accordingly, we translate utility estimates into probability estimates for the MNL model in order to compare it with RF directly.

As shown in Figure 3(a), RF, MNL, and mixed logit share a similar decreasing trend for TT\_PT, while NN presents a different decreasing pattern.
As shown in Figures 4(b)-4(d), for Wait\_Time, Rideshare, and Transfer, RF and NN also differ from two logit models. While MNL and mixed logit show a nearly linear relationship between these features and the probability of choosing \textit{PT}, the two machine-learning models reveal some nonlinear relationships. For example, the following observations can be highlighted on the RF model: 1) For TT\_PT, RF has relative flat tails before 10 minutes and after 25 minutes, showing people tend to become insensitive to very short or very long transit times; 2) travelers are more sensitive to wait times less than 5 minutes; and 3) the choice probability of \textit{PT} decreases more significantly from 0 to 1 rideshare compared to from 1 to 2 rideshares. Based on these observations, we specified piece-wise utility functions (i.e., specifying different coefficients for a variable in different data intervals) for the logit models (MNL and mixed logit). While not showing the model outputs here, we found that the model fit improved and that the coefficient estimates largely agreed with the nonlinearies revealed by the RF model. These results will be presented in a separated paper.

Therefore, partial dependence plots of machine-learning models readily reveal the nonlinearities of mode choice responses to level-of-service attributes. In contrast to the time-consuming hand-curating procedure required in logit models (often by introducing interactions terms) to reveal nonlinear relationships, machine-learning algorithms exhibit these nonlinearities automatically and thus can generate richer behavioral insights much more effectively. Machine-learning models can thus serve as an exploratory analysis tool for identifying better specifications for the logit models in order to enhance the predictive power and explanatory capabilities of logit models.



\subsubsection{Arc Elasticity and Marginal Effects}

Logit models are often applied to generate behavioral outputs such as marginal effects and elasticities to gain insights on individual travel behavior. Marginal effects (and elasticities) measure the changes of the choice probability of an alternative in response to one unit (percent) change in an independent variable. This study calculates marginal effects and arc elasticities for the level-of-service variables associated with the proposed mobility-on-demand transit system, including TT\_PT, Wait\_Time, Rideshare, and Transfer. 

The marginal effects and arc elasticity results for MNL, mixed logit, NN, and RF are presented in Table \ref{tab:marginal_effect_elas}. It is notable that we use $\Delta = 2$ min to compute the marginal effects of Wait\_Time, and we present the results for RF in two ways (all: entire market; constrained: part of the market with ``out-of-bound'' observations removed). This is mainly because the nature of the RF model: RF consists of hundreds of decision trees which apply decision rules based on discrete values, and so they may not be sensitive enough to small marginal changes and they are unable to properly predict ``out-of-bound'' observations.

Table \ref{tab:marginal_effect_elas} illustrates that the arc-elasticity and marginal-effect estimates are all negative, indicating that when the level-of-service of transit gets worse, the travelers' preferences for transit will decrease. Moreover, for Wait\_Time, Transfer, and Rideshare, the marginal-effect estimates of logit models are larger than those of NN and RF. For TT\_PT, the marginal effects and arc elasticity estimates for MNL, NN, and RF are similar in magnitude, whereas the estimates for the mixed logit model are much smaller. For RF, removing ``out-of-bound'' observations increases the marginal effects and elasticity estimates.

Therefore, we find significant differences in the behavioral outputs across the four models. Without the ground truth, it is difficult to assess the validity of these results. However, one can obtain more readily interpretable behavioral insights by converting these marginal-effect estimates into relative value-of-time measures. The following value-of-time measures are obtained by dividing all marginal effects estimates with that of transit travel time. First, the penalty of a transfer is approximately equal to 5.5 min (MNL), 13.4 min (mixed logit), 2.6 min (NN), and 3.1 min (RF) of transit travel time. Also, the penalty of a rideshare stop is equivalent to 4.2 min (MNL), 8.9 min (mixed logit), 1.0 min (NN), and 2.1 min (RF) of transit travel time. Finally, the value of one min wait time is equal to 1.5 min (MNL), 3.4 min (mixed logit), 0.2 min (NN), 0.7 min (RF) of transit travel time. The results of RF seem more realistic and more consistent with logit models.

The existing literature generally finds that the penalty effects of a transfer is larger than 5 min \citep[e.g.,][]{garcia2018transfer}, and the value of wait time is slightly larger than that of in-vehicle travel time \citep{abrantes2011meta}. Though the results of logit models seem more aligned with the empirical findings, the results of RF may still be sound. One reason for smaller penalties of RF is that TT\_PT consists of in-vehicle and out-of-vehicle travel times (the former has lower penalty compared to the latter), and thus using TT\_PT to construct the value-of-time measures may lead to smaller outputs. The other reason is that the new MOD system is expected to be app-based and highly synchronized, so passengers may perceive that the transfer will be much more convenient and they can actively wait at home after booking the trip, leading to smaller penalties for Transfer and Wait\_Time.

\begin{table}[!t]
\caption{Marginal Effects and Arc Elasticity of \textit{PT} Market Share with Respect to Transfer, Rideshare, TT\_PT, and Wait\_Time.}
\footnotesize
\begin{tabular}{cccccccc}
\hline
\multicolumn{2}{c}{\multirow{2}{*}{Variable}} & \multirow{2}{*}{$\Delta$} & \multirow{2}{*}{MNL} & \multirow{2}{*}{Mixed logit} & \multirow{2}{*}{NN} & \multicolumn{2}{c}{RF} \\ \cline{7-8} 
\multicolumn{2}{c}{}                          &                           &                      &                              &                     & All      & Constrained \\ \hline
Wait\_Time                & Marginal effects  & 1 or 2 min               & $-2.93$\%              & $-2.96$\%                      & $-0.58$\%             & $-0.83$\%  & $-1.16$\%     \\ \hline
Transfer                  & Marginal effects  & 1 unit                    & $-10.69$\%             & $-11.66$\%                     & $-6.27$\%             & $-4.60$\%  & $-5.10$\%     \\ \hline
Rideshare                 & Marginal effects  & 1 unit                    & $-8.13$\%              & $-7.74$\%                      & $-2.54$\%             & $-2.08$\%  & $-3.41$\%     \\ \hline
\multirow{2}{*}{TT\_PT}   & Marginal effects  & 1 min                    & $-1.94$\%              & $-0.87$\%                      & $-2.45$\%             & $-1.63$\%  & $-1.63$\%     \\
                          & Arc elasticity        & 10\%                      & $-0.89$                & $-0.49$                        & $-1.28$               & $-1.07$    & $-1.08$       \\ \hline
\end{tabular}
\label{tab:marginal_effect_elas}
\end{table}

\section{Discussion and Conclusion}
\label{sec7}

The increasing popularity of machine learning in transportation research raises questions regarding its advantages and disadvantages compared to conventional logit-family models used for travel behavioral analysis. The development of logit models typically focuses on parameter estimation and pays little attention to prediction (i.e., lack of a procedure to validate out-of-sample prediction accuracy). On the other hand, machine-learning models are built for prediction but are often considered as difficult to interpret and are rarely used to extract behavioral findings from the model outputs. 

This paper aims at improving the understanding of the relative strengths and weaknesses of logit models and machine learning for modeling travel mode choices. It compared logit and machine-learning models side by side using cross validation to discover their predictive and interpretability capabilities. The results showed that the best-performing machine-learning model, the RF model, significantly outperforms the logit models both at individual and aggregate levels. In fact, most machine learning models outperform the logit models. Somewhat surprsingly, the mixed logit model underperformed the MNL in terms of the out-of-sample predictive accuracy, which may result from overfitting. Moreover, to interpret the machine-learning models, we applied three techniques, including variable importance, partial dependence plots, and sensitivity analysis, to extract behavioral insights from the model outputs. 

Some of the results were illuminating. First, machine learning and logit models largely agree on variable importance and the direction of impact that each variable has on the choice outcome. However, there are some differences in the behavioral outputs (marginal effects and arc elasticities) between machine learning and logit models. Moreover, we find that the RF model can automatically capture the nonlinear effects of an independent variable on the choice outcome. This indicates that machine learning can, at minimum, serve as an exploratory analysis tool to reveal nonlinearities; researchers can then apply such information to specify logit models that can better represent behavioral preferences and have better predictive capabilities, which should be much more efficient than a hand-curating procedure typically done with statistical models.

Overall, these results are encouraging and identify many new research directions in applying machine learning to model travel behavior and forecast travel demand. Prediction and interpretation are two major topics in modeling individual choice behavior. Traditionally, each approach has focused on one aspect and ignored the other. We have demonstrated that both approaches can be applied to make predictions and infer behavior. Nonetheless, there are several major topics in travel-behavior research that we have not examined in depth. The first topic is concerned with preference heterogeneity. The development of the mixed logit model has mostly been driven by its capability to capture both observed and unobserved preference heterogeneity among individuals. We have not addressed this important topic in this paper. The second topic is on mechanisms to correct the reporting bias associated with the SP data. The SP data are generally considered as containing reporting bias due to their hypothetical nature. Logits models using joint RP and SP data have been proposed to correct for this bias \citep{train2009discrete} but, to our knowledge, no machine learning algorithms allow such a joint estimation process. 


We believe that there is great potential in merging important ideas from machine learning and logit models to develop more refined models for the research of travel behavior modeling. Besides addressing the limitations mentioned above, other possible research directions include: 1) examining which machine-learning models are more suitable than others for behavioral analysis; and 2) imposing behavioral constraints to the risk functions of machine-learning models to improve their interpretability.



\noindent 

\section*{Acknowledgements}
This research was partly funded by the Michigan Institute of Data Science (MIDAS) and by Grant 7F-30154 from the Department of Energy.



\bibliographystyle{elsarticle-harv}

 \biboptions{semicolon,round,sort,authoryear}

\bibliography{manuscript.bbl}

\begin{thebibliography}{53}
\expandafter\ifx\csname natexlab\endcsname\relax\def\natexlab#1{#1}\fi
\expandafter\ifx\csname url\endcsname\relax
  \def\url#1{\texttt{#1}}\fi
\expandafter\ifx\csname urlprefix\endcsname\relax\def\urlprefix{URL }\fi

\bibitem[{Abrantes and Wardman(2011)}]{abrantes2011meta}
Abrantes, P.~A., Wardman, M.~R., 2011. Meta-analysis of \text{UK} values of
  travel time: An update. Transportation Research Part A: Policy and Practice
  45~(1), 1--17.

\bibitem[{Ben-Akiva et~al.(1985)Ben-Akiva, Lerman, and
  Lerman}]{ben1985discrete}
Ben-Akiva, M.~E., Lerman, S.~R., Lerman, S.~R., 1985. Discrete Choice Analysis:
  Theory and Application to Travel Demand. Vol.~9. MIT press.

\bibitem[{Breiman(1996)}]{breiman1996bagging}
Breiman, L., 1996. Bagging predictors. Machine Learning 24~(2), 123--140.

\bibitem[{Breiman(2001)}]{breiman2001random}
Breiman, L., 2001. Random forests. Machine Learning 45~(1), 5--32.

\bibitem[{Breiman(2017)}]{breiman2017classification}
Breiman, L., 2017. Classification and Regression Trees. Routledge.

\bibitem[{Brownstone and Train(1998)}]{brownstone1998forecasting}
Brownstone, D., Train, K., 1998. Forecasting new product penetration with
  flexible substitution patterns. Journal of Econometrics 89~(1-2), 109--129.

\bibitem[{Cawley and Talbot(2010)}]{cawley2010over}
Cawley, G.~C., Talbot, N.~L., 2010. On over-fitting in model selection and
  subsequent selection bias in performance evaluation. Journal of Machine
  Learning Research 11~(Jul), 2079--2107.

\bibitem[{Chen et~al.(2017)Chen, Zahiri, and Zhang}]{chen2017understanding}
Chen, X.~M., Zahiri, M., Zhang, S., 2017. Understanding ridesplitting behavior
  of on-demand ride services: An ensemble learning approach. Transportation
  Research Part C: Emerging Technologies 76, 51--70.

\bibitem[{Cherchi and Cirillo(2010)}]{cherchi2010validation}
Cherchi, E., Cirillo, C., 2010. Validation and forecasts in models estimated
  from multiday travel survey. Transportation Research Record: Journal of the
  Transportation Research Board~(2175), 57--64.

\bibitem[{Christopher(2016)}]{christopher2016pattern}
Christopher, M.~B., 2016. Pattern Recognition and Machine Learning.
  Springer-Verlag New York.

\bibitem[{Farrar and Glauber(1967)}]{farrar1967multicollinearity}
Farrar, D.~E., Glauber, R.~R., 1967. Multicollinearity in regression analysis:
  The problem revisited. The Review of Economic and Statistics, 92--107.

\bibitem[{Friedman(2001)}]{friedman2001greedy}
Friedman, J.~H., 2001. Greedy function approximation: A gradient boosting
  machine. Annals of Statistics, 1189--1232.

\bibitem[{Garcia-Martinez et~al.(2018)Garcia-Martinez, Cascajo, Jara-Diaz,
  Chowdhury, and Monzon}]{garcia2018transfer}
Garcia-Martinez, A., Cascajo, R., Jara-Diaz, S.~R., Chowdhury, S., Monzon, A.,
  2018. Transfer penalties in multimodal public transport networks.
  Transportation Research Part A: Policy and Practice 114, 52--66.

\bibitem[{Gevrey et~al.(2003)Gevrey, Dimopoulos, and Lek}]{gevrey2003review}
Gevrey, M., Dimopoulos, I., Lek, S., 2003. Review and comparison of methods to
  study the contribution of variables in artificial neural network models.
  Ecological modelling 160~(3), 249--264.

\bibitem[{Goldstein et~al.(2015)Goldstein, Kapelner, Bleich, and
  Pitkin}]{goldstein2015peeking}
Goldstein, A., Kapelner, A., Bleich, J., Pitkin, E., 2015. Peeking inside the
  black box: Visualizing statistical learning with plots of individual
  conditional expectation. Journal of Computational and Graphical Statistics
  24~(1), 44--65.

\bibitem[{Golshani et~al.(2018)Golshani, Shabanpour, Mahmoudifard, Derrible,
  and Mohammadian}]{golshani2018modeling}
Golshani, N., Shabanpour, R., Mahmoudifard, S.~M., Derrible, S., Mohammadian,
  A., 2018. Modeling travel mode and timing decisions: Comparison of artificial
  neural networks and copula-based joint model. Travel Behaviour and Society
  10, 21--32.

\bibitem[{Hagenauer and Helbich(2017)}]{hagenauer2017comparative}
Hagenauer, J., Helbich, M., 2017. A comparative study of machine learning
  classifiers for modeling travel mode choice. Expert Systems with Applications
  78, 273--282.

\bibitem[{Hastie et~al.(2001)Hastie, Tibshirani, and
  Friedman}]{friedman2001elements}
Hastie, T., Tibshirani, R., Friedman, J., 2001. The Elements of Statistical
  Learning. Vol.~1. Springer series in statistics New York, NY, USA:.

\bibitem[{Hensher and Greene(2003)}]{hensher2003mixed}
Hensher, D.~A., Greene, W.~H., 2003. The mixed logit model: The state of
  practice. Transportation 30~(2), 133--176.

\bibitem[{Hensher et~al.(2005)Hensher, Rose, and Greene}]{hensher2005applied}
Hensher, D.~A., Rose, J.~M., Greene, W.~H., 2005. Applied Choice Analysis: A
  Primer. Cambridge University Press.

\bibitem[{Ho(1998)}]{ho1998random}
Ho, T.~K., 1998. The random subspace method for constructing decision forests.
  IEEE Transactions on Pattern Analysis and Machine Intelligence 20~(8),
  832--844.

\bibitem[{Hsu and Lin(2002)}]{hsu2002comparison}
Hsu, C.-W., Lin, C.-J., 2002. A comparison of methods for multiclass support
  vector machines. IEEE transactions on Neural Networks 13~(2), 415--425.

\bibitem[{Jenkins(2018)}]{RitmoTransit}
Jenkins, K., Jan 2018. New app reinvents \text{University} bus system to be
  more like \text{Uber}. The Michigan
  Daily\href{https://www.michigandaily.com/section/research/new-app-helps-turns-university-bus-system-demand-service}{.
  URL
  https://www.michigandaily.com/section/research/new-app-helps-turns-university-bus-system-demand-service}.

\bibitem[{Karlaftis and Vlahogianni(2011)}]{karlaftis2011statistical}
Karlaftis, M.~G., Vlahogianni, E.~I., 2011. Statistical methods versus neural
  networks in transportation research: Differences, similarities and some
  insights. Transportation Research Part C: Emerging Technologies 19~(3),
  387--399.

\bibitem[{Klaiber and von Haefen(2011)}]{klaiber2011random}
Klaiber, H.~A., von Haefen, R.~H., 2011. Do random coefficients and alternative
  specific constants improve policy analysis? \text{An} empirical investigation
  of model fit and prediction. Environmental and Resource Economics, 1--17.

\bibitem[{Last et~al.(2002)Last, Maimon, and Minkov}]{last2002improving}
Last, M., Maimon, O., Minkov, E., 2002. Improving stability of decision trees.
  International Journal of Pattern Recognition and Artificial Intelligence
  16~(02), 145--159.

\bibitem[{Lh{\'e}ritier et~al.(2018)Lh{\'e}ritier, Bocamazo, Delahaye, and
  Acuna-Agost}]{lheritier2018airline}
Lh{\'e}ritier, A., Bocamazo, M., Delahaye, T., Acuna-Agost, R., 2018. Airline
  itinerary choice modeling using machine learning. Journal of Choice
  Modelling.

\bibitem[{Liaw and Wiener(2002)}]{RF}
Liaw, A., Wiener, M., 2002. Classification and regression by
  \text{randomForest}. R News 2~(3), 18--22.
\newline\urlprefix\url{http://CRAN.R-project.org/doc/Rnews/}

\bibitem[{Mah{\'e}o et~al.(2017)Mah{\'e}o, Kilby, and Van~Hentenryck}]{TS2017}
Mah{\'e}o, A., Kilby, P., Van~Hentenryck, P., 2017. Benders decomposition for
  the design of a hub and shuttle public transit system. Transportation
  Science.

\bibitem[{McCallum et~al.(1998)McCallum, Nigam,
  et~al.}]{mccallum1998comparison}
McCallum, A., Nigam, K., et~al., 1998. A comparison of event models for naive
  bayes text classification. In: AAAI-98 Workshop on Learning for Text
  Categorization. Vol. 752. Citeseer, pp. 41--48.

\bibitem[{McFadden(1973)}]{mcfadden1973conditional}
McFadden, D., 1973. Conditional logit analysis of qualitative choice behaviour.
  In: Zarembka, P. (Ed.), Frontiers in Econometrics. Academic Press New York,
  New York, NY, pp. 105--142.

\bibitem[{McFadden and Train(2000)}]{mcfadden2000mixed}
McFadden, D., Train, K., 2000. Mixed \text{MNL} models for discrete response.
  Journal of applied Econometrics 15~(5), 447--470.

\bibitem[{Menard(2004)}]{menard2004six}
Menard, S., 2004. Six approaches to calculating standardized logistic
  regression coefficients. The American Statistician 58~(3), 218--223.

\bibitem[{Meyer et~al.(2017)Meyer, Dimitriadou, Hornik, Weingessel, and
  Leisch}]{e1071}
Meyer, D., Dimitriadou, E., Hornik, K., Weingessel, A., Leisch, F., 2017.
  e1071: Misc Functions of the Department of Statistics, Probability Theory
  Group (Formerly: E1071), TU Wien. \text{R} package version 1.6-8.
\newline\urlprefix\url{https://CRAN.R-project.org/package=e1071}

\bibitem[{Molnar(2018)}]{molnar2018interpretable}
Molnar, C., 2018. Interpretable Machine Learning: A Guide for Making Black Box
  Models Explainable.

\bibitem[{Mullainathan and Spiess(2017)}]{mullainathan2017machine}
Mullainathan, S., Spiess, J., 2017. Machine learning: An applied econometric
  approach. Journal of Economic Perspectives 31~(2), 87--106.

\bibitem[{Murdoch et~al.(2019)Murdoch, Singh, Kumbier, Abbasi-Asl, and
  Yu}]{murdoch2019interpretable}
Murdoch, W.~J., Singh, C., Kumbier, K., Abbasi-Asl, R., Yu, B., 2019.
  Interpretable machine learning: definitions, methods, and applications. arXiv
  preprint arXiv:1901.04592.

\bibitem[{Omrani(2015)}]{omrani2015predicting}
Omrani, H., 2015. Predicting travel mode of individuals by machine learning.
  Transportation Research Procedia 10, 840--849.

\bibitem[{Omrani et~al.(2013)Omrani, Charif, Gerber, Awasthi, and
  Trigano}]{omrani2013prediction}
Omrani, H., Charif, O., Gerber, P., Awasthi, A., Trigano, P., 2013. Prediction
  of individual travel mode with evidential neural network model.
  Transportation Research Record: Journal of the Transportation Research Board
  2399~(1), 1--8.

\bibitem[{Quinlan(2014)}]{quinlan2014c4}
Quinlan, J.~R., 2014. C4. 5: Programs for Machine Learning. Elsevier.

\bibitem[{Ridgeway(2017)}]{gbm}
Ridgeway, G., 2017. gbm: Generalized Boosted Regression Models. \text{R}
  package version 2.1.3.
\newline\urlprefix\url{https://CRAN.R-project.org/package=gbm}

\bibitem[{Ripley(2016)}]{tree}
Ripley, B., 2016. tree: Classification and Regression Trees. \text{R} package
  version 1.0-37.
\newline\urlprefix\url{https://CRAN.R-project.org/package=tree}

\bibitem[{Train(2009)}]{train2009discrete}
Train, K.~E., 2009. Discrete choice methods with simulation. Cambridge
  University Press.

\bibitem[{Venables and Ripley(2002)}]{stats}
Venables, W.~N., Ripley, B.~D., 2002. Modern Applied Statistics with S, 4th
  Edition. Springer, New York, iSBN 0-387-95457-0.
\newline\urlprefix\url{http://www.stats.ox.ac.uk/pub/MASS4}

\bibitem[{Wager and Athey(2018)}]{wager2018estimation}
Wager, S., Athey, S., 2018. Estimation and inference of heterogeneous treatment
  effects using random forests. Journal of the American Statistical Association
  113~(523), 1228--1242.

\bibitem[{Wang and Ross(2018)}]{wang2018machine}
Wang, F., Ross, C.~L., 2018. Machine learning travel mode choices: Comparing
  the performance of an extreme gradient boosting model with a multinomial
  logit model. Transportation Research Record: Journal of the Transportation
  Research Board.

\bibitem[{Wong et~al.(2018)Wong, Farooq, and Bilodeau}]{wong2018discriminative}
Wong, M., Farooq, B., Bilodeau, G.-A., 2018. Discriminative conditional
  restricted boltzmann machine for discrete choice and latent variable
  modelling. Journal of Choice Modelling 29, 152--168.

\bibitem[{Xie et~al.(2003)Xie, Lu, and Parkany}]{xie2003work}
Xie, C., Lu, J., Parkany, E., 2003. Work travel mode choice modeling with data
  mining: decision trees and neural networks. Transportation Research Record:
  Journal of the Transportation Research Board~(1854), 50--61.

\bibitem[{Yan et~al.(2018)Yan, Levine, and Zhao}]{YAN2018}
Yan, X., Levine, J., Zhao, X., 2018. Integrating ridesourcing services with
  public transit: An evaluation of traveler responses combining revealed and
  stated preference data. Transportation Research Part C: Emerging
  Technologies.
\newline\urlprefix\url{http://www.sciencedirect.com/science/article/pii/S0968090X18310398}

\bibitem[{Zhang(2004)}]{zhang2004optimality}
Zhang, H., 2004. The optimality of naive \text{Bayes}. AA 1~(2), 3.

\bibitem[{Zhang and Xie(2008)}]{zhang2008travel}
Zhang, Y., Xie, Y., 2008. Travel mode choice modeling with support vector
  machines. Transportation Research Record: Journal of the Transportation
  Research Board~(2076), 141--150.

\bibitem[{Zhao and Hastie(2017)}]{zhao2017causal}
Zhao, Q., Hastie, T., 2017. Causal interpretations of black-box models. Journal
  of Business \& Economic Statistics, to appear.

\bibitem[{Zhao et~al.(2019)Zhao, Yan, and Van~Hentenryck}]{zhao2019modeling}
Zhao, X., Yan, X., Van~Hentenryck, P., 2019. Modeling heterogeneity in
  mode-switching behavior under a mobility-on-demand transit system: An
  interpretable machine learning approach. arXiv preprint arXiv:1902.02904.

\end{thebibliography}

\end{document}